%% file: main_DAC.tex
\documentclass[sigconf,9pt,nonacm]{acmart}

\input{packages}

\title{\methodname: Hard-constrained Differentiable Co-Exploration Method for Neural Architectures and Hardware Accelerators}
\title{Enabling Hard Constraints in Differentiable Neural Network and Accelerator Co-Exploration \\
\vspace{2mm}
    \large{ \normalfont Deokki Hong\textsuperscript{1}, Kanghyun Choi\textsuperscript{1}, Hye Yoon Lee\textsuperscript{1}, Joonsang Yu\textsuperscript{2}, Noseong Park\textsuperscript{1}, Youngsok Kim\textsuperscript{1}, and Jinho Lee\textsuperscript{1*}} \\
    \normalsize \normalfont 
    \textsuperscript{1}College of Computing, Yonsei University,
    \textsuperscript{2}CLOVA ImageVision, CLOVA AI Lab, NAVER \\
    \textsuperscript{1} \{dk.hong, kanghyun.choi, hylee817, noseong, youngsok, leejinho\}@yonsei.ac.kr \hspace{2mm}
    \textsuperscript{2} joonsang.yu@navercorp.com
}

\thanks{* Corresponding author}

\renewcommand{\shortauthors}{Hong, et al.}
\renewcommand{\shorttitle}{Enabling Hard Constraints in Differentiable Neural Network and Accelerator Co-Exploration}

\begin{document}


\begin{abstract}
Co-exploration of an optimal neural architecture and its hardware accelerator is an approach of rising interest which addresses the computational cost problem, especially in low-profile systems. 
The large co-exploration space is often handled by adopting the idea of differentiable neural architecture search. 
However, despite the superior search efficiency of the differentiable co-exploration, it faces a critical challenge of not being able to systematically satisfy hard constraints such as frame rate.
To handle the hard constraint problem of differentiable co-exploration, we propose \methodname,  
which searches for hard-constrained solutions without compromising the global design objectives.
By manipulating the gradients in the interest of the given hard constraint, high-quality solutions satisfying the constraint can be obtained.
\end{abstract}

\maketitle
\vspace{-1mm}
\section{Introduction}
The primary interest of most \emph{Deep Neural Network} (DNN) researches has been the application performance (i.e., accuracy).
However, it also led to the rapid growth in the network size that require immense computational resources for execution.
In recent years, numerous schemes have appeared to mitigate the resource problem, mostly belonging to one of these categories -- \emph{network-side optimization} and \emph{hardware-side optimization}.
Network-side optimization refers to refining the architecture of a neural network to reduce computations while maintaining a comparable accuracy~\citep{mobilenetv2, dorefa}.  
Hardware-side optimization often involves improving DNN execution efficiency by using optimized hardware, also known as \emph{accelerator}s~\citep{eyeriss, tpu}.
Unfortunately, effort from one side often hinders the benefits coming from the other.
For example, the main advantage of depth-wise separable convolution operation used in MobileNet~\citep{mobilenetv2} comes from its structure which uses a single channel for its operation.
However, Google's TPU~\citep{tpu}, a renowned accelerator, mainly utilizes channel-level parallelism for gaining speedup. In consequence, MobileNet results in a poor execution time on TPUs~\citep{edgetpunas}.

Co-exploration of hardware accelerator and network architecture~\citep{li2020edd, fu2021autonba, dance, abdelfattah2020best, hao2019fpgadnn, lu2019neural, yang2020coexploration} is therefore a natural direction to fulfill both goals of accuracy and hardware metrics such as latency, energy consumption and silicon chip area.
To address the large search space of the co-exploration, 
\emph{differentiable co-exploration} methods~\citep{dance, li2020edd, fu2021autonba} are considered as promising approaches due to their ability to quickly explore the search space compared to its reinforcement learning based counterparts~\citep{abdelfattah2020best, hao2019fpgadnn, lu2019neural, yang2020coexploration, lin2021naas}.

Unfortunately, differentiable co-exploration has a serious drawback of being unable to deal with hard constraints that are critical in many real-world scenarios. 
For instance, an important constraint of object detection system~\citep{yolo} is to meet the frame rate of the camera (e.g., 30 frames per second). 
In addition, a mobile subsystem running on a limited battery often has a power budget. 
Because differentiable co-exploration methods rely on a single loss function, they often fail to satisfy the constraints, and have to blindly undergo a several repetitions of hyper-parameter tuning and re-exploration.

\sloppypar{ 
In order to address the problem, we present \emph
{\methodname} 
(\underline{\textbf{H}}ard-constrained \underline{\textbf{D}}ifferentiable
e\underline{\textbf{X}}ploration),  
which enables hard-constrained differentiable co-exploration of neural architecture and hardware accelerator. 
%
}
The key concept of our proposal 
is a gradient manipulation method which ensures that the solution does not drift away from meeting the constraints.
In addition to the gradients from the global loss, 
we calculate the gradient from the hardware constraints, which is used to manipulate gradient of the global loss, if any constraint violations, such that i) the dot product of the two are positive (i.e., they point to a similar direction), and ii) the direction can alleviate the violation of the constraints. 
%

To the best of our knowledge, this is the first work that considers hard constraints in a differentiable co-exploration problem. 
We conduct an extensive amount of evaluation to demonstrate that \methodname can 1) satisfy the hardware constraints even under tight constraints and 2) search solutions without compromising the quality. 

Our contributions can be summarized as follows.
\begin{itemize}
    \item We propose \methodname, a hard-constrained differentiable co-exploration method for network and accelerator in order to find valid solutions without trial-and-errors.
    \item We propose using gradient manipulation to gradually move solutions towards the constraint-satisfying region.
    \item We provide an extensive evaluation for \methodname to show its constraint-meeting capability and efficiency. 
\end{itemize}

\vspace{-2mm}
\section{Background and Related Work}
\vspace{-1mm}
\subsection{Neural Architecture Search}
\vspace{-1mm}
Neural Architecture Search (NAS) refers to the technique of automating the neural network design process.
To mitigate its huge cost (thousands of GPU-hours), Differentiable NAS~\citep{liu2019darts} has been proposed as an efficient alternative, which converts the problem to training a supernet that reduces the time cost by a few digits.

Regardless, optimizing solely on network performance is insufficient as they do not take hardware efficiency into account. 
Some recent works~\citep{cai2019proxylessnas} that address this issue consider hardware costs by adding loss terms, and  
some attempt to reduce the network size for latency constraints using simple latency models~\citep{nayman2021hardcorenas,aows}.
However, they only consider the network design, and cannot be used for co-exploration since the relation between network accuracy and hardware structure is not reflected in the model. 

\subsection{DNN-Accelerator Co-exploration} 
The early work on the co-exploration utilize variants of reinforcement learning, or evolutionary algorithm to leverage its simplicity~\cite{hao2019fpgadnn,lu2019neural,jiang2019accuracy,abdelfattah2020best, yang2020coexploration,lin2021naas}. 
Each candidate network is trained for evaluation, while the accelerator design is analyzed for hardware efficiency.
These values create rewards used by the agent to create the next candidate solution. 
%
%
However, they all inherit the same problem from RL-based NAS methods in which they require expensive training to evaluate each candidate solution. 
To worsen the matter, co-exploration requires even larger network/hardware search space than searching only for networks. 

In such regard, differentiable approaches were adopted to co-exploration~\citep{li2020edd}.
Auto-NBA~\citep{fu2021autonba} used a differentiable accelerator search engine to build a joint-search pipeline, and 
DANCE~\citep{dance} trained auxiliary neural networks for hardware search and cost evaluation. 
%
However, none of the above properly addresses the hard constraint problem. 
In this work, we propose a holistic method of handling hard constraints on differentiable co-exploration.

\section{Motivational Experiment}
\label{sec:difficulty}

\begin{figure}[t]
    \centering
    \includegraphics[width=0.95\columnwidth]{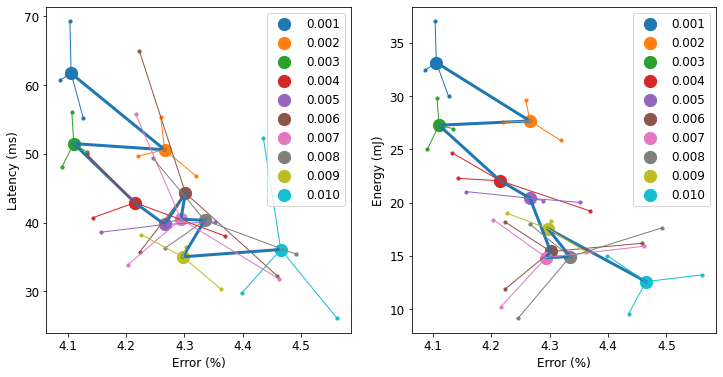} 
    \vspace{-3.5mm}
    \caption{A motivational experiment. In each plot, we swept the value $\lambda_{Cost}$ 0.001 to 0.010. 
    It is clear that the trajectory is not strictly linear to $\lambda_{cost}$ with high variations. }
    \label{fig:motivational exp}
    \vspace{-3mm}
\end{figure}




The most straightforward and na\"ive way to handle hard constraints within differentiable co-exploration would be to tune the relative weight to the hardware cost.
For example, below is the loss function used in differentiable co-exploration~\citep{dance,fu2021autonba}.
\begin{align}
    \mathcal{L}oss &= \mathcal{L}oss_{CE} + \lambda_{Cost}Cost_{HW},
    \label{eq:danceloss}
\end{align}
which is designed co-optimize accuracy and hardware cost simultaneously, and $\lambda_{Cost}$ balances the two terms\footnote{It is different from hyperparameters of typical machine learning formulation where the two terms serve toward a single objective.}.
By increasing $\lambda_{Cost}$, one can indirectly instruct the search process to consider hardware metrics more. 
However, giving a larger penalty 
does not directly lead to reduction in the value of a constrained metric. 
\figurename~\ref{fig:motivational exp} plots how changing $\lambda_{Cost}$ in \cref{eq:danceloss} from 0.001 to 0.010 affects the latency/energy and the classification error for CIFAR-10 dataset. 
Searches were done three times for each setting and plotted with the same colors with large dots for their averages.
Even though some trend is observed that depends on $\lambda_{Cost}$, inconsistency in both direction and variance of the trajectory is more dominant. 

Consider a scenario where a designer wants to design a neural network-accelerator architecture pair with latency under some constraint (e.g., \SI{33.3}{\milli\second}), using the conventional co-exploration methods.
The designer would try searching with some initial $\lambda_{Cost}$ and try adjusting the value over the course of multiple searches.
However, such inconsistency between $\lambda_{Cost}$ and the latency makes it extremely difficult to obtain the adequate solution, not to mention the huge time cost of performing the search numerous times.
Despite the difficulties that lie in tackling a hard-constrained co-exploration problem,
designing an effective strategy is necessary.

\section{Hard-Constrained Co-exploration}
\subsection{Problem Definition}
The mathematical formulation of hard-constrained differentiable co-exploration is as below:
\begin{align}
\small
    \argmin_{\alpha, \beta} &(\mathcal{L}oss_{NAS} (w^*, net(\alpha)) + \lambda_{Cost} Cost_{HW}(eval(\alpha, \beta))), \notag \\
    &\textrm{s.t.}\ \ w^* = \argmin_{w} (\mathcal{L}oss_{NAS}(w, net(\alpha))), \ \ \ t\leq T,
\label{eq:object}
\end{align}
where $t$ denotes the current value of constrained metric such as latency or energy, and $T$ is the target value (e.g., 33.3 \SI{}{\milli\second} for latency).
%
%
$\alpha$ and $\beta$ denote network architecture parameters and hardware accelerator configuration, respectively. ${w}$ is the weights of the NAS supernet and ${net(\alpha)}$ is the current dominant network architecture selected. 
${eval(\alpha, \beta)}$ indicates the hardware metrics evaluated for $\alpha$ and $\beta$. 
The objective of co-exploration is expressed using two distinct evaluation metrics, which are neural architecture loss ($\mathcal{L}oss_{NAS}$) and hardware cost ($Cost_{HW}$) defined from the user. 

\subsection{Differentiable Co-exploration Framework}
\begin{figure}[t]
    \centering
    \includegraphics[width=.4\textwidth]{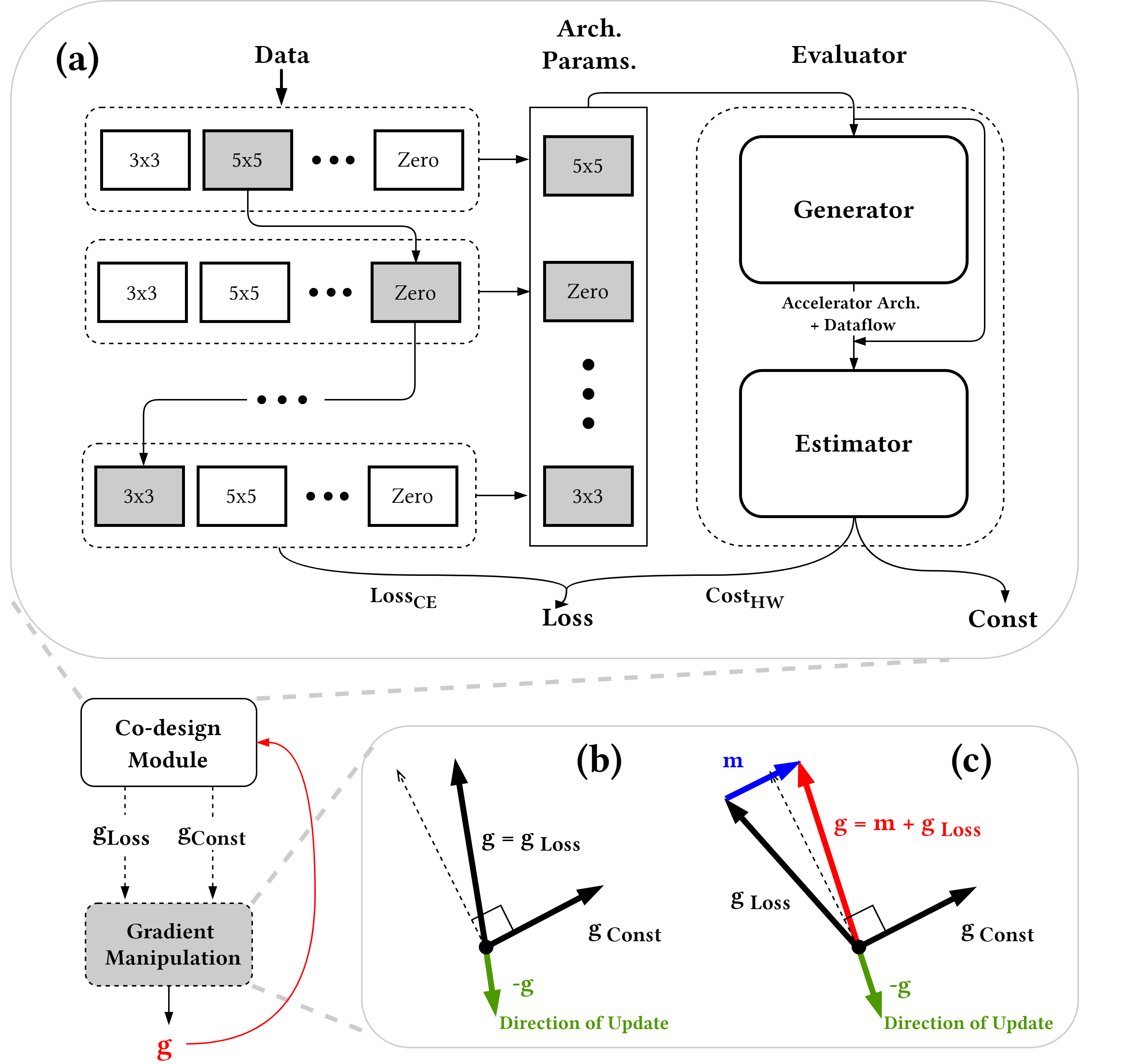} 
    \vspace{-3mm}
    \caption{Overall structure of HDX. 
    }\vspace{-4mm}
    \label{fig:HDX}
\end{figure}

Although our main contribution is that we enable hard constraints, we explain our framework for the co-exploration since they are closely related.
\figurename~\ref{fig:HDX} illustrates the overall architecture of the proposed method, being similar to existing methods~\cite{dance, fu2021autonba}. 
\figurename~\ref{fig:HDX} (a) is the network search module. 
This module searches for network architecture by choosing a path from the 
supernet. 
The network structure is then fed to the evaluator module.


The evaluator network $eval()$ is the key to the differentiable co-exploration that enables the gradient to flow into the supernet, considering the relation between the hardware accelerator.
It is a composition of two subnetworks: a hardware generator $gen()$ and an estimator $est()$.
The hardware generator takes the neural architecture parameters as inputs and uses them to output the optimal hardware implementation ($\beta$ from \eq{object}). 
It is jointly trained during the co-exploration so that the generator does not depend on certain cost function, and can adapt to the constraint. 
The estimator network outputs the hardware-related metrics by taking output of the generator and the network.
It is pre-trained according to the network and the accelerator search space.
For pre-training the estimator, traditional (non-differentiable) cost estimation frameworks such as MAESTRO~\citep{maestro_toppicks2020}, Timeloop~\citep{TimeLoop}, and Accelergy~\citep{iccad_2019_accelergy} are used as ground truth. 
After pre-training, the estimator is frozen during the exploration and is only used to infer the hardware cost given a network architecture. 
With these, we convert \eq{object} as below:
\begin{align}
\small
    \argmin_{\alpha} (\mathcal{L}oss_{NAS} &(w^*, net(\alpha)) + \lambda_{Cost} Cost_{HW}(est(\alpha, gen(v^*, \alpha)))), \notag \\
    \textrm{s.t.}\ \ w^* &= \argmin_{w} (\mathcal{L}oss_{NAS}(w, net(\alpha))), \notag \\
                 \ \ v^* &= \argmin_{v} (Cost_{HW}(est(\alpha, gen(v, \alpha)))),
\label{eq:cosearch}
\end{align}
where $v$ is the weights for the hardware generator.

\subsection{Enabling Hard-Constraints with Gradient Manipulation}
In addition to the differentiable co-exploration methodology, we suggest the novel idea of gradient manipulation as an effective solution to the hard constraint problem. 
Direct manipulation of gradients is a strategy often used in achieving multiple goals,
such as in continual learning~\citep{saha2021gradient} or differential equations~\citep{dpm}. 
In this paper, we present a solution to apply gradient manipulation to the co-exploration problem in the interest of satisfying hard constraints.

The diagrams on \figurename~\ref{fig:HDX} (b) and (c) show a high-level abstraction of our gradient manipulation method. 
The main idea is to artificially generate a force that can push the gradient in the direction that \emph{agree}s with the constraint. 
The conditions under which the method is applied to compute the new gradient $g$ are defined as below:
\begin{equation}
    g = 
        \begin{cases}
            g_{\mathcal{L}oss} & \text{, if } t \leq T \\
            &\text{ or } t > T \wedge g_{\mathcal{L}oss} \cdot g_{\mathcal{C}onst} \geq \text{0,} \\
            m_{\alpha} + g_{\mathcal{L}oss} &  \text{, otherwise} \\
        \end{cases}
\end{equation}
\begin{equation}
    g_{\mathcal{C}onst} = \frac{\partial \max(t - T ,0)}{\partial \alpha}.
\end{equation}

In the above equation, $g_{\mathcal{L}oss}$ is the original gradient from the global loss function defined as
\begin{equation}
\label{eq:global_loss}
    \mathcal{L}oss = \mathcal{L}oss_{NAS} + \lambda_{Cost}\cdot Cost_{HW},
\end{equation}
as in \eq{cosearch}, and $g_{\mathcal{C}onst}$ is the gradient of constraint loss that we define as: $\mathcal{C}onst=max(t - T ,0)$. 
\JL{I moved the def. of t and T to section 3} 
Note that $t$ is a function of $\alpha$, and thus can be backpropagated to find the gradient with respect to $\alpha$.
In an ideal case where the $t$ $\leq$ $T$, the constraint is already met so we do nothing. 
In the unfortunate case when the constraint is still not met, we calculate for the dot product of the two gradients to determine the agreement in their directions. 
If $g_{\mathcal{L}oss} \cdot g_{\mathcal{C}onst} \text{$\geq$ 0}$ (i.e., the angle between two gradients is less than 90\textdegree), it means gradient descent update will contribute towards satisfying the constraint.
Thus it is interpreted as an agreement in direction and the same $g_{\mathcal{L}oss}$ is used unmodified. 
\figurename~\ref{fig:HDX} (b) depicts this scenario. 
However, if they disagree as illustrated in \figurename~\ref{fig:HDX} (c) (i.e., $g_{\mathcal{L}oss} \cdot g_{\mathcal{C}onst} \text{$<$ 0}$), we force the gradient to shift its direction by $m_{\alpha}$, which is obtained from $(m_{\alpha} + g_{\mathcal{L}oss}) \cdot g_{\mathcal{C}onst} \text{$\geq$ 0}$ to guarantee decrease in target cost after gradient descent. 
It can be reformulated as $m_{\alpha} \cdot g_{\mathcal{C}onst} + g_{\mathcal{L}oss} \cdot g_{\mathcal{C}onst} = \delta$ where $\delta \text{$\geq$ 0}$ is a small value for ensuring gradual movement towards satisfying the constraint. 
%

For updating $\alpha$ and $w$, 
we solve for optimal $m_{\alpha}$ with respect to $\alpha$, which are the parameters for the network architecture. 
To minimize the effect of $m_{\alpha}$ on $g_{\mathcal{L}oss}$, we use a pseudoinverse-based solution that is known to minimize the size of $||m_{\alpha}||_2^2$ as below:  
\begin{equation}
\label{eq:optimal_m}
    m_\alpha^* = \frac{-(g_{\mathcal{L}oss} \cdot g_{\mathcal{C}onst})  + \delta}{||g_{\mathcal{C}onst}||_2^2} g_{\mathcal{C}onst}.
\end{equation}
%
%
%
In order to control the magnitude of the pull, we use a small multiplying factor $p>0$ on $\delta$. The policy for updating $\delta$ using $p$ is as follows: Some initial value $\delta_{0}$ exists for $\delta$. If the target metric fails to meet the constraint, $\delta$ is multiplied by $1+p$ to strengthen the pull ($\delta^{'} = (1+p)\delta$). In the other case when the constraint is satisfied, $\delta$ is reset to its initial value ($\delta^{'} = \delta_{0}$).

Note that we also train $v$, weights for the hardware generator using gradient descent. 
Thus we compute for $m_v^*$ in the same manner, but use $g_{\mathcal{C}ost_{HW}}$ in place of $g_{\mathcal{L}oss}$ for updating the generator.

Although a single constraint is already a challenging target, our method can be further generalized to accommodate multiple constraints.
Now the gradient is modified only in the direction of individual constraints that do not comply. We provide a more generalized formulation:
\begin{equation}
    g = 
        \begin{cases}
            g_{\mathcal{L}oss} & \text{, if } \bigwedge_{i=1}^{n}(t_{i} \leq T_{i}) \\
            &\text{ or } \bigvee_{i=1}^{n}(t_i > T_{i}) \wedge g_{\mathcal{L}oss} \cdot g_{\mathcal{C}onst} \geq \text{0,} \\
            m_{\alpha} + g_{\mathcal{L}oss} &  \text{, otherwise} 
        \end{cases}
\end{equation}
\begin{equation}
    g_{\mathcal{C}onst} = \frac{\partial \sum_{i=1}^{n}\max(t_{i} - T_{i} ,0)}{\partial \alpha}.
\end{equation}


\subsection{Implementation Details}
\textbf{\ \ \ \ \ Hardware cost function.}
In this work, we choose the inference latency, energy, and the chip area as the widely used hardware metrics. 
Considering all of them, a commonly used cost function is multiplying them (i.e., EDP, EDAP) as in \cite{dance, fu2021autonba}. 
However, we found that the energy is usually easier to optimize for, and using such cost function unfairly favors energy-oriented designs.
Therefore, we use a balanced weighted sum for the cost function as below.
\begin{equation}
\small
    Cost_{HW} = C_EEnergy + C_LLatency + C_AArea.
\vspace{-1mm}  
\label{eq:cost_hw}
\end{equation}

\textbf{Estimator and Generator Network.} Following \cite{dance}, we model both the estimator and generator with five-layer Multi-Layer Perceptron (MLP) with residual connections. 
To train the estimator, we first build a dataset by randomly sampling 10.8M network-accelerator pairs ($2.95\mathrm{e}{-9}$ \% of the total search space) from our search space which are evaluated on hardware metrics using Timeloop~\citep{TimeLoop} and Accelergy~\citep{iccad_2019_accelergy}. 
Using this dataset, the estimator is trained for 200 epochs with the batch size of 256. The weight update is done using Adam optimizer with the learning rate of 1e-4. 
The accuracy of the estimator was over 99\% for all metrics, being powerful enough as an engine for co-exploration.
The generator is randomly initialized and jointly trained with the NAS supernet. 
As the manipulated gradient from the hard-constraint is back-propagated, the generator learns to create accelerators that comply with the constraint on given neural network architecture.

\textbf{Search Space.}
We use ProxylessNAS \citep{cai2019proxylessnas} as a NAS backbone with path sampling to train $\alpha$.
It consists of multiple settings of MBConv operation with kernel size \{3, 5, 7\} and expand ratio \{3, 6\}. 
The total number of layers is 18 and 21 for CIFAR-10~\cite{cifar} and ImageNet~\cite{imagenet} dataset, respectively. 
However, our method is orthogonal to the NAS implementation and has the flexibility to choose from any differentiable NAS algorithms, such as DARTS~\cite{liu2019darts} or OFA~\cite{ofa}. 

We use Eyeriss~\citep{eyeriss} as the accelerator's backbone architecture. 
It is composed of a two-dimensional Processing Element~(PE) array where 
each PEs has a Multiply-Accumulate~(MAC) unit attached to a register file. 
Therefore, hardware accelerator design space comprises PE array size from 12$\times$8 to 20$\times$24, register file size per PE from 16B to 256B.
In addition, the search space includes dataflow of Weight-Stationary (WS) similar to \citep{tpu}, Output-Stationary (OS) similar to \citep{shidiannao} and Row-Stationary (RS) similar to \citep{eyeriss}.

\section{Experiments}
\subsection{Experimental Environment}
We have conducted experiments on \methodname using CIFAR-10~\citep{cifar} and ImageNet~\citep{imagenet} dataset. 
%
%
%
For all the hardware metrics (latency, energy, and chip area) reported, we have used the direct evaluation on the designed hardware from Timeloop~\cite{TimeLoop} and Accelergy~\cite{iccad_2019_accelergy} instead of the values outputted by the estimator to avoid any possible error in the learned model.
Evaluation of network accuracy is done by training the final network architecture, which we train from scratch for 300 epochs using the batch size of 64.
We use SGD optimizer with Nesterov momentum \citep{nesterov}, and cosine learning rate scheduling with 0.008 as its initial value, while weight decay term and momentum is 1e-3 and 0.9 respectively. Augmentation for the train data is adopted from AutoAugment \citep{Cubuk2019AutoAugmentLA}, on both dataset.

\subsection{Comparison with Existing Methods}
\label{sec:comparison}
\tablename~\ref{tbl:comparison} lists the existing differentiable co-exploration methods in comparison with \methodname.
We compare the following methods:
%
\begin{itemize}[leftmargin=*]   
    \item \textbf{NAS $\rightarrow$ HW}: A plain differentiable NAS~\cite{cai2019proxylessnas} only, followed by an exhaustive hardware search using Timeloop~\cite{TimeLoop}.

    \item \textbf{Auto-NBA~\cite{fu2021autonba}}: A differentiable co-exploration method that directly searches for hardware parameters using gradient descent, where the relation between hardware and the DNN is expressed as a lookup table.

    \item \textbf{DANCE~\cite{dance}}: State-of-the-art differentiable co-exploration method without hard constraints 
    where the generator and estimator are modeled as neural networks. 

    \item \textbf{DANCE + Soft Constraint}: To represent a reasonable approach for considering constraints with existing solutions, we have added a soft-constraint term $\lambda_{Soft} \cdot max(t/T-1,0)$ as in \citep{hu2020tf}. 
    
    \item \textbf{\methodname}: The proposed method with $p=1\mathrm{e}{-2}$.
       
\end{itemize}
As displayed in \tablename~\ref{tbl:comparison}, \methodname is the only method that considers hard constraints with differentiable co-exploration.

For a quantitative comparison, we devised an algorithm for finding constrained solution using co-exploration without hard-constraints. 
Because a constrained metric (e.g., latency) being too low often means an inefficient solution (in terms of other metrics (e.g., accuracy and energy), we set the criteria of having a solution of 50\%\textasciitilde 100\% of the target constraint.
A rough sketch of the algorithm similar to binary search is given below:
\begin{enumerate}
    \item Choose a control parameter ($\lambda_{Soft}$ and $\lambda_{Cost}$) that indirectly affects the metric under constraint (e.g., latency). 
    \item Perform a search using the default control parameter value.
    \item Perform searches by doubling the control parameter each step until the metric is under the constraint.
    \item If the metric is under 50\% of the target value, shrink the control parameter in a binary search manner. 
\end{enumerate}
Note that the problem cannot exactly be solved by binary search, because of huge per-search variations and non-linear relations as demonstrated in Section~\ref{sec:difficulty}.
In the actual implementation, more features are added to deal with corner cases and to avoid falling into wrong parameter region (which can happen from per-search variations).
We ran the algorithm 100 times and report the average.
We also report the average error as a proxy to the solution quality.

As presented in \tablename~\ref{tbl:comparison}, all baselines require multiple searches to find constraint-satisfying solutions. 
For DANCE, it takes 6.6 searches with 12.2 GPU-hours on average and 4.9 searches and 9.1 GPU-hours on even with soft constraints.
On the other hand, \methodname always searches for constraint-satisfying solutions in a single search.
Furthermore, the number of searches for baselines strongly depend on the choice of the default value of control parameter.
Guessing a right initial parameter is difficult, and could result in even larger repetition cost. 
This advocates the need for \methodname.

\begin{table}[t]
\vspace{-1mm}
    \centering
     \resizebox{\columnwidth}{!}
    {
    \begin{tabular}{lcccccccc}
     \toprule
  \multirow{2}{*}{Method} &  \multirow{2}{*}{\makecell{Hard \\ Constraint}} &  \multirow{2}{*}{\makecell{NN-HW \\Relation}} & \multicolumn{3}{c}{Search with 60 FPS Constraint}  \\ 
  \cmidrule(lr){4-6}
                          &                                   &                                  & \#Searches  & Cost*  & Avg. Err. (\%) \\        
    \midrule
        NAS~\cite{cai2019proxylessnas} $\rightarrow$ HW search & \xmark & \xmark & 5.0 & 10.9h & 7.26 \\
        Auto-NBA~\cite{fu2021autonba} & \xmark & \cmark & 6.8 & 10.2h  & 5.67 \\
        DANCE~\cite{dance} & \xmark & \cmark & 6.6 & 12.2h & 5.32 \\
        DANCE + Soft const.~\cite{hu2020tf} & \xmark & \cmark & 4.9 & 9.1h  & 5.36 \\

        \midrule
       \textbf{ \methodname (Proposed)} & \cmark & \cmark &  \textbf{1} & \textbf{2.0h} &  \textbf{4.65} \\
        \bottomrule
        \multicolumn{6}{r}{*GPU-hours}

    \end{tabular}
    }\\
   
\caption{Comparison of Differentiable Co-explorations}
\label{tbl:comparison}
\vspace{-10mm}
\end{table}

\subsection{Co-exploration Results}
\label{sec:results}

\figurename~\ref{fig:masterplot} plots the co-exploration experimental results from multiple techniques on CIFAR-10 dataset.
In the experiments, we have set two different constraints for the latency: 16.6\SI{}{\milli\second} (60 fps) and 33.3\SI{}{\milli\second} (30 fps).  
For all co-exploration methods, we obtain five solutions by varying $\lambda_{Cost}$ from 0.001 to 0.005 to fairly compare the various design points obtained by each approach.
For $NAS\rightarrow HW$, we obtain 10 solutions of various range for reference, because directly applying $\lambda_{Cost}$ is infeasible.
For DANCE and Auto-NBA, the black markers are unconstrained, and colored markers are obtained with soft constraint of the corresponding colors.

\input{figs/master_three}

\input{figs/table_quality}

In all experiments, we have used $C_E=2.9$, $C_L=6.2$, and $C_A=1.0$ from \eq{cost_hw}, which makes the difference scale of each metric approximately the same to make a fair comparison. 
For the NAS only method, $Cost_{HW}$ was used only for the hardware design phase.

\figurename~\ref{fig:masterplot} (left) and (mid) show the relation between error and latency. 
The colored horizontal bars represent the two latency targets we applied.
It can be easily seen that all solutions found by \methodname satisfy the given hard constraints regardless of the value of $\lambda_{Cost}$. 
Furthermore, all solutions have the latency right below the constraint, showing that the solutions did not over-optimize for the constrained metric (latency).
%
DANCE~\citep{dance} and Auto-NBA~\cite{fu2021autonba} were able to exploit the trade-off between hardware metrics and accuracy, but has no control over meeting the constraint. 
Even with soft-constraint terms, they mostly failed to obtain in-constraint solutions. 
Auto-NBA at a glance seems to be slightly better at meeting the constraints, but it is because its baseline method favors hardware-efficient solutions over high-accuracy ones, not because of its ability to meet the constraint, exemplified by the fact that there is no solution with high accuracy, or latency under \SI{16.6}{\milli\second}.

\subsection{Solution Quality Found by \methodname}
\label{sec:quality}
In this subsection, we demonstrate that \methodname can 1) handle constraints from all three metrics (latency, energy, and chip area), 2) handle multiple constraints, and 3) obtain solutions of good overall quality.
\figurename~\ref{fig:masterplot} (right) plots $Cost_{HW}$ and error together, which allows evaluating quality of the solutions in terms of Pareto-optimality.
Because \figurename~\ref{fig:masterplot} (left) and (mid) overlook the other metrics, comparing the $Cost_{HW}$ together is required to be fair. 

From the plot, it is clear that 
quality of solutions from \methodname is better than the NAS$\rightarrow$HW method, and has no degradation from the existing co-exploration methods. 
In fact, the tightly constrained (16.6\SI{}{\milli\second}) solutions even find better solution than those of the existing solutions in terms of Pareto-optimality.

To further study the quality of the solutions found by \methodname, we have conducted another set of experiments.
We selected a few solutions found from DANCE method as `Anchor' solutions and listed them in \tablename~\ref{tab:quality}. 
From those, we chose either one or all three of the hardware metrics to be fixed as the hard constraint, and performed co-explorations using \methodname.
Because it is guaranteed that such solution exists, a good method should be able to find a solution meeting the constraint, of at least a similar quality.
As in the Section~\ref{sec:results}, all of the 8 cases we have examined succeeded in finding a valid solution.
%
Furthermore, all the solutions show similar global loss values from the anchor solutions as shown in the rightmost column.

\vspace{-1mm}
\subsection{Results from ImageNet Dataset}
\vspace{-1mm}

\begin{table}[b]
\vspace{-3mm}
    \centering
    \scriptsize
    \caption{Experimental Results for ImageNet}
\vspace{-3mm}
    \label{tab:imgnet}
    \begin{tabular}{lccccc}
    \toprule

Method	&	in-const?	&	Lat. (ms)	&	Error (\%)	&	CostHW	&	Loss	\\
\midrule											
\multirow{2}{*}{NAS$\rightarrow$HW}	&	\xmark	&	242.92	&	24.84	&	46.29	&	1.99	\\
	&	\xmark	&	135.39	&	28.83	&	24.26	&	2.17	\\
\midrule											
\multirow{2}{*}{DANCE}	&	\xmark	&	165.98	&	25.46	&	29.37	&	2.04	\\
	&	\cmark	&	121.58	&	25.28	&	25.92	&	2.09	\\
\midrule											
\multirow{2}{*}{DANCE+Soft Const.}	&	\xmark	&	188.69	&	25.09	&	33.14	&	1.99	\\

	&	\cmark	&	105.65	&	26.37	&	25.58	&	2.08	\\
\midrule

\multirow{2}{*}{\textbf{HDX (Proposed)}}	&	\cmark	&	92.06	&	25.01	&	24.48	&	1.98	\\

	&	\cmark	&	112.11	&	25.20	&	22.63	&	2.00	\\

\bottomrule
    \end{tabular}
\end{table}

\input{figs/sense}

\tablename~\ref{tab:imgnet} shows the co-exploration results from ImageNet dataset~\citep{imagenet}, under 125 ms constraint.
As displayed in the table, \methodname always succeeded in finding a solution within constraint where the others often failed to satisfy.
Furthermore, the Top-1 error and the global loss shows that the quality of the solution found by \methodname is not compromised at all, compared to DANCE or its variant.


\vspace{-1mm}
\subsection{Sensitivity Study on Pulling Magnitude}
\vspace{-1mm}

In \methodname, the only hyperparameter is $p$ that controls the pulling magnitude.
\figurename~\ref{fig:sense} illustrates how the global loss and latency changes over latency-constrained (33.3 \SI{}{\milli\second}) explorations, with varying $p$ of 1e-2, 7e-3, and 4e-3.

Regardless of the value of $p$, the curve for the constrained value shows a similar trend. 
At the beginning, the global loss becomes mainly optimized, while the latency stays steady. 
During this phase the pulling magnitude $\delta$ (See \eq{optimal_m}) is still growing, and is not strong enough to make meaningful changes.
At certain point, 
$\delta$ becomes strong enough to pull the solution towards lowering latency.
When the latency satisfies the constraint, 
global loss starts to decrease while maintaining the latency. 
There is no significant discrepancy between the final solution in the global loss and the latency, which shows that \methodname is insensitive to the hyperparameter $p$.


\subsection{Analysis on the Searched Solutions}
\begin{figure}
\begin{subfigure}[t]{0.4\columnwidth}
\includegraphics[width=\textwidth, trim={.3cm .1cm 1cm .1cm}, clip=true]{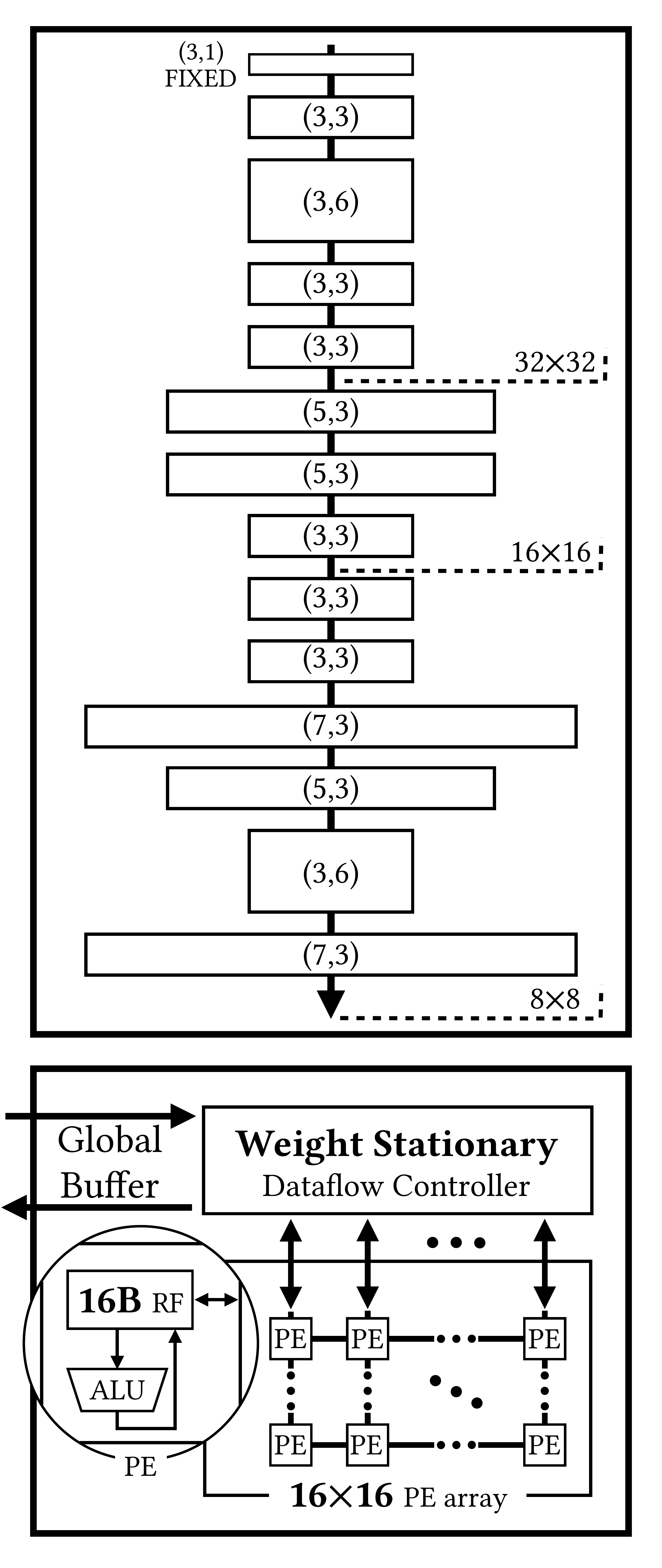} 
\vspace{-6.5mm}
\caption{}\label{fig:nn_hw_60fps}
\end{subfigure}
\hspace{3mm}
\begin{subfigure}[t]{0.4\columnwidth}
\includegraphics[width=\textwidth, trim={.3cm .1cm 1cm .1cm}, clip=true]{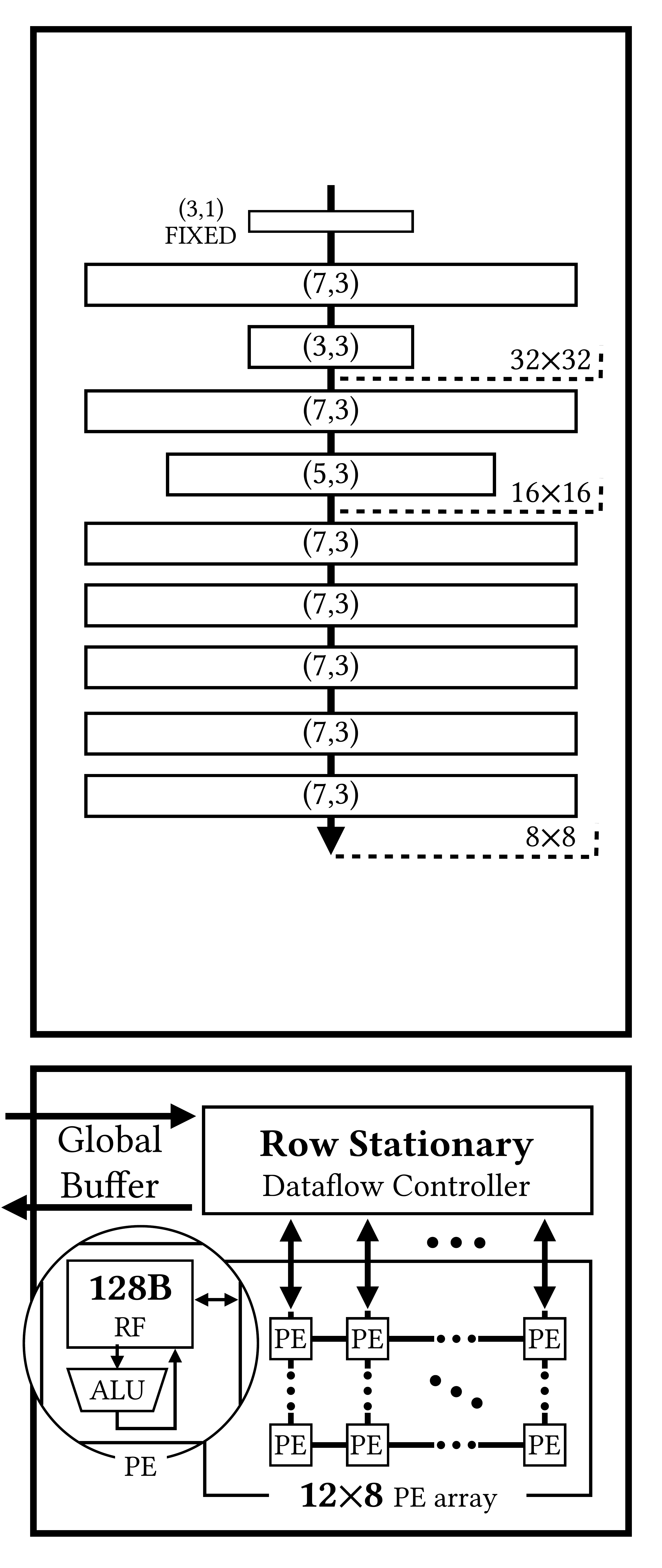} 
\vspace{-6.5mm}
\caption{}\label{fig:nn_hw_60fps}
\end{subfigure}
\vspace{-3mm}
\caption{Searched network and accelerator for 60 fps and 30 fps constraints. ($m$,$n$) refers MBConv block with $m\times m$ kernel and $n$ expand ratio.}
\vspace{-5mm}
\label{fig:nn_hw_pair}
\end{figure}

\cref{fig:nn_hw_pair} visualizes the network and accelerator searched for 60 fps (a) and 30 fps (b) constraints.
For the found design pair (a), the design contains relatively smaller kernels, more layers, and a powerful accelerator. 
To meet a tight constraint while maintaining accuracy, the network has small kernels, mainly of $3\times3$.  
Using smaller kernels quadratically reduces the number of multiplications. 
Therefore, decreasing the kernel size and increasing number of layers is a good choice for reducing inference latency.
Looking at the accelerator design, it has relatively large PE array (16$\times16$) to achieve low latency.
It takes weight stationary (WS) dataflow, which is known to have low latency. 
In addition, there are some kernels with high channel expand ratio in the network. 
WS exploits channel parallelism for fast execution, and thus has advantage over the found network.

On the other hand, in the design for 30 fps (b), the design settles at a solution that can optimize the energy consumption while satisfying the constraint.
The design uses larger kernels in the network and row stationary (RS) dataflow in the accelerator.
RS is known to have good energy efficiency~\cite{eyeriss}, and exploits parallelism from spatial dimensions of kernel and the activation. 
Thus, having larger kernels have advantages on RS dataflow.
To reduce the energy consumption, the design has fewer PEs (12$\times$8), larger RFs to save off-chip access energy, and fewer layers in the network. 



\vspace{-1mm}
\section{Conclusion}
\vspace{-1mm}
In this paper, we proposed \methodname, a hard-constrained differentiable co-exploration method. 
By conditionally applying gradient manipulation that moves the solution towards meeting the constraints, 
hard constraints can be reliably satisfied with high-quality solutions.
We believe this proposal would ease the development of DNN based systems by a significant amount.


\begin{acks}
This work has been supported by the National Research Foundation of Korea (NRF) grant funded by the Korea government (MSIT) (2022R1C1C1008131,  
2022R1C1C1011307),  
and 
Institute of Information \& communications Technology Planning \& Evaluation
(IITP) grant funded by the Korea government (MSIT) 
(2020-0-01361,
Artificial Intelligence Graduate School Program (Yonsei University)
).
\end{acks}



\bibliographystyle{ACM-Reference-Format}
\bibliography{refs}


\balance

\end{document}

%% file: packages.tex
\input{math_commands.tex}

\usepackage{hyperref}
\usepackage{url}
\usepackage{multirow}
\usepackage{graphicx}
\usepackage{xspace}
\usepackage{subcaption}
\usepackage{verbatim} 
\usepackage{booktabs} 

\usepackage{multirow}
\usepackage{array, makecell} %
\usepackage{siunitx}
\usepackage{alphalph} 
\usepackage{tablefootnote}

\usepackage{enumitem}

\usepackage[export]{adjustbox} 

\usepackage[capitalize]{cleveref}
\crefname{section}{Sec.}{Secs.}
\Crefname{section}{Section}{Sections}
\Crefname{table}{Table}{Tables}
\crefname{table}{Table}{Tables}

\definecolor{olivegreen}{rgb}{0, 0.6, 0}
\newcommand{\JL}[1]{{\color{olivegreen}[\textbf{\sc JLee}: \textit{#1}]}}

\newcommand{\KH}[1]{{\color{purple}[\textbf{\sc KH}: \textit{#1}]}}
\newcommand{\HY}[1]{{\color{blue}[\textbf{\sc HY}: \textit{#1}]}}
\newcommand{\DK}[1]{{\color{teal}[\textbf{\sc DK}: \textit{#1}]}}

\renewcommand{\JL}[1]{}
\renewcommand{\KH}[1]{}
\renewcommand{\HY}[1]{}
\renewcommand{\DK}[1]{}

\newcommand{\methodname}{HDX\xspace}

\newcommand{\eq}[1]{Eq.~\ref{eq:#1}\xspace}

\usepackage{pgfplots}
\usepackage{pgfplotstable}
\usepgfplotslibrary{groupplots}
\usepackage{tikz}
\usetikzlibrary{patterns}

\usepackage{pifont}
\newcommand{\cmark}{\color{olivegreen}\ding{51}}%
\newcommand{\xmark}{\color{red}\ding{55}}%

%% file: math_commands.tex
\usepackage{amsmath,amsfonts,bm}

\def\eqref#1{equation~\ref{#1}}

\def\1{\bm{1}}

\DeclareMathAlphabet{\mathsfit}{\encodingdefault}{\sfdefault}{m}{sl}
\SetMathAlphabet{\mathsfit}{bold}{\encodingdefault}{\sfdefault}{bx}{n}

\DeclareMathOperator*{\argmin}{arg\,min}

%% file: figs/master_three.tex
   \newcommand{\thirdwidth}{0.42\columnwidth}
   \newcommand{\yshwidth}{-1.7em}

\begin{figure}[t]

\centering
    \begin{tikzpicture}
\input{figs/data/data_cifar_dac}
\pgfplotsset{
compat=1.11,
legend image code/.code={
\draw[mark repeat=2,mark phase=2]
plot coordinates {
(0cm,0cm)
(0.15cm,0cm)        
(0.3cm,0cm)         
};%
}
}                     
\begin{groupplot}[group style={vertical sep=2.8em,horizontal sep=2.5em,group size= 3 by 1},height=4cm,width=0.3\columnwidth]


\nextgroupplot[
width=\thirdwidth,
font=\footnotesize,
title style={yshift=-1.5mm},
xmajorticks=true,
ymax = 80,
ymin = 0,
ytick={0,20,...,120},
xmin=3.8,
xmax=5.5,
xlabel = {Error (\%)},
xlabel shift=-.5em,
xlabel near ticks,
xlabel style={at={(ticklabel cs:0.5)},font =\scriptsize},
x tick label style = {font =\scriptsize, text width = 1.4cm, align = center, anchor = north},
ylabel={Latency (\SI{}{\milli\second})},
ylabel near ticks,
ylabel shift=\yshwidth,
ylabel style = {at={(ticklabel cs:0.5)},font=\scriptsize},
y tick label style = {font =\scriptsize, text width = .6cm, align = right, anchor = east,
 /pgf/number format/fixed
},
	legend style={fill=white, at={(0.95,0.95)}, font=\tiny,
	anchor=north east,legend columns=4,
	/tikz/every column/.append style={column sep=55pt},
	transpose legend},
]

\addplot [mark=none,thick,color=red,forget plot]table[x=Error,y=sixty]{\bars};
\addplot [mark=none,thick,color=blue,forget plot]table[x=Error,y=thirty]{\bars};

\addplot [mark=+,only marks,thick,color=red,forget plot]table[x=Error,y=Latency]{\concodesixty};

\addplot [mark=+,only marks,thick,color=blue,forget plot]table[x=Error,y=Latency]{\concodethirty};



\nextgroupplot[
width=\thirdwidth,
font=\footnotesize,
title style={yshift=-1.5mm},
xmajorticks=true,
ymax = 80,
ymin = 0,
ytick={0,20,...,120},
xmin=3.8,
xmax=5.5,
xlabel = {Error (\%)},
xlabel shift=-.5em,
xlabel near ticks,
xlabel style={at={(ticklabel cs:0.5)},font =\scriptsize},
x tick label style = {font =\scriptsize, text width = 1.4cm, align = center, anchor = north},
ylabel={Latency (\SI{}{\milli\second})},
ylabel near ticks,
ylabel shift=\yshwidth,
ylabel style = {at={(ticklabel cs:0.5)},font=\scriptsize},
y tick label style = {font =\scriptsize, text width = .6cm, align = right, anchor = east,
 /pgf/number format/fixed
},
	legend style={fill=white, at={(0.95,0.95)}, font=\scriptsize,
	anchor=north east,legend columns=4,
	/tikz/every column/.append style={column sep=55pt},
	},
	legend to name=masterlegend,
  legend style={legend columns=4,
  },
]
]

\addplot [mark=+,only marks,thick,color=black]table[x=Error,y=Latency]{\dummy};
\addlegendentry{\methodname}

\addplot [mark=none,thick,color=red,forget plot]table[x=Error,y=sixty]{\bars};
\addplot [mark=none,thick,color=blue,forget plot]table[x=Error,y=thirty]{\bars};

\addplot [mark=x,mark options={},only marks,color=black]table[x=Error,y=Latency]{\flops};
\addlegendentry{NAS $\rightarrow$ HW}

\addplot [mark=triangle,only marks,color=black]table[x=Error,y=Latency]{\dummy};
\addlegendentry{Auto-NBA~\cite{fu2021autonba}}

\addplot [mark=diamond,thick,only marks,color=black]table[x=Error,y=Latency]{\dance};
\addlegendentry{DANCE~\cite{dance}}

\addplot [mark=diamond,thick,only marks,color=black, forget plot]table[x=Error,y=Latency]{\dummy};

\addplot [mark=none,thick,fill=white]table[x=Error,y=Latency]{\dummy};
\addlegendentry{Latency Const.}

\addplot [mark=square*, only marks,thick,color=red]table[x=Error,y=Latency]{\dummy};
\addlegendentry{16.6 \SI{}{\milli\second} (60 fps)}
\addplot [mark=square*, only marks,thick,color=blue, fill=blue]table[x=Error,y=Latency]{\dummy};
\addlegendentry{33.3 \SI{}{\milli\second} (30 fps)}


\addplot [mark=diamond,only marks,thick,color=red,forget plot]table[x=Error,y=Latency]{\linearsixty};

\addplot [mark=diamond,only marks,thick,color=blue,forget plot]table[x=Error,y=Latency]{\linearthirty};

\addplot [mark=triangle,only marks,thick,color=red,forget plot]table[x=Error,y=Latency]{\nbasixty};

\addplot [mark=triangle,only marks,thick,color=blue,forget plot]table[x=Error,y=Latency]{\nbathirty};



\nextgroupplot[
width=\thirdwidth,
font=\footnotesize,
title style={yshift=-1.5mm},
xmajorticks=true,
ymax = 30,
ymin = 0,
ytick={0,5,10,15,20,...,120},
xmin=3.8,
xmax=5.5,
xlabel = {Error (\%)},
xlabel shift=-.5em,
xlabel near ticks,
xlabel style={at={(ticklabel cs:0.5)},font =\scriptsize},
x tick label style = {font =\scriptsize, text width = 1.4cm, align = center, anchor = north},
ylabel={$Cost_{HW}$},
ylabel near ticks,
ylabel shift=\yshwidth,
ylabel style = {at={(ticklabel cs:0.5)},font=\scriptsize},
y tick label style = {font =\scriptsize, text width = .6cm, align = right, anchor = east,
 /pgf/number format/fixed
},
	legend style={fill=white, at={(0.95,0.95)}, font=\tiny,
	anchor=north east,legend columns=4,
	/tikz/every column/.append style={column sep=55pt},
	transpose legend},
]

\addplot [mark=+,only marks,thick,color=black]table[x=Error,y=HWCost]{\dummy};

\addplot [mark=diamond,only marks,thick,color=black]table[x=Error,y=HWCost]{\dummy};

\addplot [mark=diamond,thick,only marks,color=black]table[x=Error,y=HWCost]{\dance};

\addplot [mark=triangle,only marks,thick,color=red]table[x=Error,y=HWCost]{\nbathirty};

\addplot [mark=triangle,only marks,thick,color=blue]table[x=Error,y=HWCost]{\nbasixty};

\addplot [mark=x,mark options={fill=lightgray},only marks,color=black]table[x=Error,y=HWCost]{\flops};

\addplot [mark=diamond,only marks,color=red,thick,forget plot]table[x=Error,y=HWCost]{\linearsixty};

\addplot [mark=diamond,only marks,color=blue,thick,forget plot]table[x=Error,y=HWCost]{\linearthirty};


\addplot [mark=+,only marks,thick,color=red,forget plot]table[x=Error,y=HWCost]{\concodesixty};

\addplot [mark=+,only marks,thick,color=blue,forget plot]table[x=Error,y=HWCost]{\concodethirty};


\end{groupplot}

jdd  \node[right=1em,inner sep=0pt] at (rel axis cs:-2.15,1.25) {\pgfplotslegendfromname{masterlegend}};
\end{tikzpicture}
\vspace{-7mm}
\caption{Co-exploration results. (left) and (mid) represent the latency and (right) represent the hardware cost.
Colored marks are methods with constraints of the same color. }
\vspace{-3mm}
\label{fig:masterplot}
\end{figure}

%% file: figs/data/data_cifar_dac.tex
    
\pgfplotstableread{
Error Latency HWCost
4.10	69.23	2.18E+01
4.22	49.65	1.67E+01
4.11	56.11	1.81E+01
4.13	49.92	1.53E+01
4.35	40.12	1.27E+01
}\dance

\pgfplotstableread{
Error Latency HWCost
4.45	28.95	10.91532137
4.69	24.29	8.570141113
5.04	24.30	8.556119463
4.61	32.90	12.08707796
4.77	29.76	10.10901835
5.23	24.25	8.52518177
4.58	27.67	9.842209016
4.46	29.63	10.03190101
4.57	27.51	9.457146288
4.45	32.02	11.64877336
4.87	25.62	8.898752028
5.23	25.26	8.844910921
}\nba

\pgfplotstableread{
Error Latency HWCost
4.45	28.95	10.91532137
4.69	24.29	8.570141113
5.04	24.30	8.556119463
4.61	32.90	12.08707796
4.77	29.76	10.10901835
5.23	24.25	8.52518177
}\nbathirty

\pgfplotstableread{
Error Latency HWCost
4.58	27.67	9.842209016
4.56	29.63	10.03190101
4.59	27.51	9.457146288
4.46	32.02	11.64877336
4.96	25.62	8.898752028
5.23	25.26	8.844910921
}\nbasixty
    
\pgfplotstableread{
Error Latency HWCost
4.01	102.86	2.19E+01
4.12	63.34	2.04E+01
4.19	61.90	1.88E+01
4.02	115.15	2.50E+01
3.92	109.71	2.37E+01
4.06	70.44	2.21E+01
11.03	7.28	3.09E+00
7.07	16.44	4.88E+00
5.75	23.01	6.02E+00
5.37	25.23	6.41E+00
4.72	33.19	1.08E+01
4.35	42.99	1.47E+01
4.01	102.86	2.19E+01
4.12	63.34	2.04E+01
4.19	61.90	1.88E+01
4.02	115.15	2.50E+01
3.92	109.71	2.37E+01
4.06	70.44	2.21E+01
11.03	7.28	3.09E+00
7.07	16.44	4.88E+00
5.75	23.01	6.02E+00
5.37	25.23	6.41E+00
4.72	33.19	1.08E+01
4.35	42.99	1.47E+01
4.01	102.86	2.19E+01
4.12	63.34	2.04E+01
4.19	61.90	1.88E+01
4.02	115.15	2.50E+01
3.92	109.71	2.37E+01
4.06	70.44	2.21E+01
11.03	7.28	3.09E+00
7.07	16.44	4.88E+00
5.75	23.01	6.02E+00
5.37	25.23	6.41E+00
4.72	33.19	1.08E+01
4.35	42.99	1.47E+01
4.196	6.32E+01	2.02E+01
4.24	5.23E+01	1.61E+01
4.412	4.18E+01	1.31E+01
5.533	2.31E+01	6.05E+00
7.255	1.60E+01	4.73E+00
}\flops
    
\pgfplotstableread{
Error Latency HWCost
4.15	50.84	1.68E+01
4.35	42.49	1.46E+01
4.21	46.53	1.44E+01
4.35	31.82	1.07E+01
4.43	35.48	1.15E+01
}\linearsixty

\pgfplotstableread{
Error Latency HWCost
4.10	58.91	1.84E+01
4.03	58.26	1.77E+01
4.14	51.78	1.70E+01
4.35	37.36	1.22E+01
4.36	32.11	1.15E+01
}\linearthirty
\pgfplotstableread{
Error Latency HWCost
4.14	61.66	1.98E+01
4.07	50.48	1.68E+01
4.32	53.76	1.66E+01
4.26	50.68	1.68E+01
4.26	36.94	1.32E+01
}\linearfifteen

\pgfplotstableread{
Error Latency HWCost
4.860	13.39	5.59E+00
4.532	15.63	6.12E+00
4.522	15.19	6.09E+00
4.674	16.13	6.49E+00
6.71	8.11E+00	4.13E+00
}\concodesixty
\pgfplotstableread{
Error Latency HWCost
4.741	30.78	1.12E+01
4.167	31.09	1.05E+01
4.472	32.55	1.07E+01
4.300	28.92	9.76E+00
4.323	3.22E+01	1.26E+01
}\concodethirty
\pgfplotstableread{
Error Latency HWCost
4.900	53.68	1.63E+01
4.800	59.53	1.95E+01
4.700	59.77	1.94E+01
4.600	48.38	1.62E+01
}\concodefifteen

\pgfplotstableread{
Error   sixty thirty fifteen
0   16.67 33.33 66.67
10   16.67 33.3 66.67
}\bars

\pgfplotstableread{
Error   Latency HWCost
0   0   0
}\dummy

%% file: figs/table_quality.tex
\begin{table}[b]
\setlength{\tabcolsep}{2pt}
\vspace{-3mm}
    \centering
    \footnotesize
    \caption{Results Showing the Quality of Solutions \vspace{-3mm}}
    \label{tab:quality}
    \begin{tabular}{clcccccc}
    \toprule
         Index & Constrained & Lat. (\SI{}{\milli\second}) & E (\SI{}{\milli\joule}) & Area ($\SI{}{\milli\meter\squared}$)  & Error (\%) & $Cost_{HW}$ &  Loss \\
\midrule
\multirow{5}{*}{A}  & Anchor  & \textcolor{blue}{69.23} & \textcolor{olivegreen}{37.00} & \textcolor{orange}{2.53} & 4.10 $\pm$ 0.16 & 21.84 & 0.632 \\
\cmidrule(lr){2-8}
                    & Latency           & \textbf{\textcolor{blue}{43.99}} & 21.79 & 2.10 & 4.20 $\pm$ 0.07 & 13.87 & 0.624 \\
                    & Energy            & 51.98 & \textbf{\textcolor{olivegreen}{29.18}} & 2.53 & 4.38 $\pm$ 0.17 & 17.44 & 0.630 \\
                    & Chip Area         & 64.00 & 34.82 & \textbf{\textcolor{orange}{2.53}} & 4.05 $\pm$ 0.06 & 20.56 & 0.629 \\
                    & All               & \textbf{\textcolor{blue}{63.72}} & \textbf{\textcolor{olivegreen}{12.09}} & \textbf{\textcolor{orange}{1.86}} & 4.12 $\pm$ 0.18 & 13.29 & 0.623 \\
\midrule
\multirow{5}{*}{B}  & Anchor  & \textcolor{blue}{49.65} & \textcolor{olivegreen}{27.53} & \textcolor{orange}{2.53} & 4.22 $\pm$ 0.06 & 16.67 & 0.638 \\
\cmidrule(lr){2-8}
                    & Latency           & \textbf{\textcolor{blue}{48.02}} & 27.33 & 2.53 & 4.27 $\pm$ 0.09 & 16.41 & 0.644 \\
                    & Energy            & 95.02 & \textbf{\textcolor{olivegreen}{24.45}} & 1.89 & 4.05 $\pm$ 0.10 & 20.76 & 0.648 \\
                    & Chip Area         & 54.74 & 29.81 & \textbf{\textcolor{orange}{2.53}} & 4.11 $\pm$ 0.13 & 17.96 & 0.645 \\
                    & All               & \textbf{\textcolor{blue}{41.32}} &  {\color{white}0}\textbf{\textcolor{olivegreen}{8.59}} & \textbf{\textcolor{orange}{1.86}} & 4.35 $\pm$ 0.05 &  {\color{white}0}9.50 & 0.629 \\

        \bottomrule
        \multicolumn{8}{r}{*Bold colored numbers indicate that they are}  \\
        \multicolumn{8}{r}{under constraint of the same colored non-bold numbers.}
        \vspace{-3mm}
    \end{tabular}
\end{table}

%% file: figs/sense.tex
   
\begin{figure}[t]
\centering
\hspace{-1cm}
\captionsetup[subfigure]{oneside,margin={.8cm,0cm}}
\begin{subfigure}{.30\columnwidth}
\centering
\pgfplotsset{
compat=1.11,
legend image code/.code={
\draw[mark repeat=2,mark phase=2]
plot coordinates {
(0cm,0cm)
(0.15cm,0cm)        
(0.3cm,0cm)         
};%
}
}                               
    \begin{tikzpicture}
\input{figs/data/sensedata}
\begin{axis}[
axis y line*=right, 
width=1.4\columnwidth, height=3.5cm,
xmajorticks=true,
xmin=-10,
xmax=130,
ymajorgrids,
ymax=200,
ymin = 0,
xlabel = {Epoch},
xlabel shift=-.5em,
xlabel near ticks,
xlabel style={at={(ticklabel cs:0.95)},font =\rmfamily\scriptsize},
x tick label style = {font =\scriptsize, text width = 1.4cm, align = center, anchor = north},
ylabel near ticks,
ylabel style = {font=\scriptsize, yshift=+.1cm},
y tick label style = {font =\scriptsize, anchor = east, 
 /pgf/number format/fixed, xshift=.6cm
},
ymajorticks=false,
legend cell align={left},
legend style={draw=none, fill=none, at={(0.647,.58), font=\tiny},
anchor=south,legend columns=1,
/tikz/every even column/.append style={column sep=0.5cm}},
	]  

\addplot[thick,color=olivegreen] table [x=epoch, 
     y expr=\thisrow{latency}*50] {\ten};\addlegendentry{Latency}

\addplot[thick,color=blue] table [x=epoch, y=latency] {\dummy};\addlegendentry{Loss}
\addplot[mark=none,color=red] table [x=epoch, y expr=\thisrow{latency}*50] {\constraint};

\end{axis}

\begin{axis}[
width=1.4\columnwidth, height=3.5cm,
xmajorticks=false,
xmin=-10,
xmax=130,
ymax = 3,
ymin = 0,
ytick={0,1,2},
ylabel={Global Loss},
ylabel near ticks,
ylabel style = {font=\scriptsize, yshift=-.1cm},
y tick label style = {font =\scriptsize, anchor = east, 
 /pgf/number format/fixed,
},
legend cell align={left},
legend style={draw=none, fill=none, at={(0.6,.55), font=\tiny},
anchor=south,legend columns=2,
/tikz/every even column/.append style={column sep=0.5cm}},
]  

\addplot[thick, color=blue] table [x=epoch, y=loss] {\ten};

\end{axis}

\end{tikzpicture}  
\vspace{-5mm}  
\caption{\vspace{-3mm}$p$=1e-2}

\end{subfigure}
\begin{subfigure}{.30\columnwidth}
\centering
\pgfplotsset{
compat=1.11,
legend image code/.code={
\draw[mark repeat=2,mark phase=2]
plot coordinates {
(0cm,0cm)
(0.15cm,0cm)        
(0.3cm,0cm)         
};%
}
}                               
    \begin{tikzpicture}
\input{figs/data/sensedata}
\begin{axis}[
axis y line*=right, 
width=1.4\columnwidth, height=3.5cm,
xmajorticks=true,
xmin=-10,
xmax=130,
ymajorgrids,
ymax=200,
ymin = 0,
xlabel = {Epoch},
xlabel shift=-.5em,
xlabel near ticks,
xlabel style={at={(ticklabel cs:0.95)},font =\rmfamily\scriptsize},
x tick label style = {font =\scriptsize, text width = 1.4cm, align = center, anchor = north},
ymajorticks=false,
ylabel style = {font=\scriptsize, yshift=+.1cm},
y tick label style = {font =\scriptsize, anchor = east, 
 /pgf/number format/fixed, xshift=.6cm
},
legend cell align={left},
legend style={draw, fill=white, at={(0.647,.55), font=\tiny},
anchor=south,legend columns=1,
/tikz/every even column/.append style={column sep=0.5cm}},
	]  

\addplot[thick,color=olivegreen] table [x=epoch, 
     y expr=\thisrow{latency}*50] {\seven};

\addplot[thick,color=blue] table [x=epoch, y=latency] {\dummy};
\addplot[mark=none,color=red] table [x=epoch, y expr=\thisrow{latency}*50] {\constraint};
\end{axis}

\begin{axis}[
width=1.4\columnwidth, height=3.5cm,
xmajorticks=false,
xmin=-10,
xmax=130,
ymax = 3,
ymin = 0,
ymajorticks=false,
ylabel style = {font=\scriptsize, yshift=-.1cm},
y tick label style = {font =\scriptsize, anchor = east, 
 /pgf/number format/fixed,
},
legend cell align={left},
legend style={draw=none, fill=none, at={(0.6,.55), font=\tiny},
anchor=south,legend columns=2,
/tikz/every even column/.append style={column sep=0.5cm}},
]  

\addplot[thick, color=blue] table [x=epoch, y=loss] {\seven};

\end{axis}

\end{tikzpicture}  
\vspace{-5mm}  
\caption{\vspace{-3mm}$p$=7e-3}

\end{subfigure}
\begin{subfigure}{.30\columnwidth}
\centering
\pgfplotsset{
compat=1.11,
legend image code/.code={
\draw[mark repeat=2,mark phase=2]
plot coordinates {
(0cm,0cm)
(0.15cm,0cm)        
(0.3cm,0cm)         
};%
}
}                               
    \begin{tikzpicture}
\input{figs/data/sensedata}
\begin{axis}[
axis y line*=right, 
width=1.4\columnwidth, height=3.5cm,
xmajorticks=true,
ymajorgrids,
ymax=200,
ymin = 0,
xmin=-10,
xmax=130,
ytick={0,50,100,150},
xlabel = {Epoch},
xlabel shift=-.5em,
xlabel near ticks,
xlabel style={at={(ticklabel cs:0.95)},font =\rmfamily\scriptsize},
x tick label style = {font =\scriptsize, text width = 1.4cm, align = center, anchor = north},
ylabel = {Latency (ms)},
ylabel style = {font=\scriptsize, yshift=+.1cm},
y tick label style = {font =\scriptsize, anchor = east, 
 /pgf/number format/fixed, xshift=.6cm
},
legend cell align={left},
legend style={draw, fill=white, at={(0.647,.55), font=\tiny},
anchor=south,legend columns=1,
/tikz/every even column/.append style={column sep=0.5cm}},
	]  

\addplot[thick,color=olivegreen] table [x=epoch, 
     y expr=\thisrow{latency}*50] {\four};

\addplot[thick,color=blue] table [x=epoch, y=latency] {\dummy};

\addplot[mark=none,color=red] table [x=epoch, y expr=\thisrow{latency}*50] {\constraint};

\end{axis}

\begin{axis}[
width=1.4\columnwidth, height=3.5cm,
xmajorticks=false,
xmin=-10,
xmax=130,
ymax = 3,
ymin = 0,
ymajorticks=false,
ylabel style = {font=\scriptsize, yshift=-.1cm},
y tick label style = {font =\scriptsize, anchor = east, 
 /pgf/number format/fixed,
},
legend cell align={left},
legend style={draw=none, fill=none, at={(0.6,.55), font=\tiny},
anchor=south,legend columns=2,
/tikz/every even column/.append style={column sep=0.5cm}},
]  

\addplot[thick, color=blue] table [x=epoch, y=loss] {\four};

\end{axis}

\end{tikzpicture}  
\vspace{-5mm}  
\caption{\vspace{-3mm}$p$=4e-3}

\end{subfigure}
\caption{Sensitivity to $p$ on \methodname. The red lines represents latency constraint at 33.3 ms.}
\vspace{-4mm}
\label{fig:sense}
\end{figure}

%% file: figs/data/sensedata.tex
\pgfplotstableread{
epoch	loss	latency	hw
1	1.6093	1.7784309	30.485004
2	1.6074	2.7493377	46.770054
3	1.6112	3.091916	50.056435
4	1.6077	3.1399508	50.736706
5	1.585	3.1010032	50.265396
6	1.5878	3.0906775	50.9511
7	1.5795	3.075262	49.96289
8	1.5781	2.6240377	48.454025
9	1.5636	2.5810184	45.776943
10	1.5421	2.589581	47.679783
11	1.5508	3.0735633	49.182636
12	1.5517	3.0735633	49.182636
13	1.5405	2.589581	47.679783
14	1.5257	2.589581	47.679783
15	1.5244	2.589581	47.679783
16	1.5108	2.6017442	47.769627
17	1.5023	2.5987713	47.84092
18	1.4925	2.5987713	47.84092
19	1.4952	2.5987713	47.84092
20	1.4855	2.5987713	47.84092
21	1.4739	2.6021593	47.989002
22	1.457	2.6021593	47.989002
23	1.4588	2.5987713	47.84092
24	1.4325	2.6021593	47.989002
25	1.4294	2.601562	48.061695
26	1.4503	2.5987713	47.84092
27	1.424	2.5987713	47.84092
28	1.418	2.5987713	47.84092
29	1.3935	2.6056876	48.057926
30	1.404	2.6056876	48.057926
31	1.3903	2.590584	47.645473
32	1.37	2.590584	47.645473
33	1.3651	2.57541	46.91261
34	1.3521	2.533668	45.41202
35	1.3481	2.52346	45.0369
36	1.3849	1.8548356	27.077328
37	1.3943	1.5660362	21.083511
38	1.4389	1.2849666	17.646175
39	1.4303	1.0406508	14.607554
40	1.4907	0.7067742	10.420017
41	1.4709	0.65971535	9.816524
42	1.4424	0.65971535	9.816524
43	1.4374	0.65971535	9.816524
44	1.4245	0.65971535	9.816524
45	1.4287	0.65971535	9.816524
46	1.4279	0.65971535	9.816524
47	1.4097	0.65971535	9.816524
48	1.4051	0.65971535	9.816524
49	1.4002	0.65971535	9.816524
50	1.3897	0.65971535	9.816524
51	1.3866	0.65971535	9.816524
52	1.3881	0.65971535	9.816524
53	1.3746	0.65971535	9.816524
54	1.3761	0.65971535	9.816524
55	1.3562	0.65971535	9.816524
56	1.3609	0.65971535	9.816524
57	1.3516	0.65971535	9.816524
58	1.349	0.65971535	9.816524
59	1.3485	0.65971535	9.816524
60	1.3409	0.65971535	9.816524
61	1.34	0.65971535	9.816524
62	1.3377	0.65971535	9.816524
63	1.3272	0.65971535	9.816524
64	1.3306	0.65971535	9.816524
65	1.3249	0.65971535	9.816524
66	1.319	0.65971535	9.816524
67	1.3273	0.65971535	9.816524
68	1.3136	0.65971535	9.816524
69	1.3161	0.65971535	9.816524
70	1.3101	0.65971535	9.816524
71	1.3059	0.65971535	9.816524
72	1.3042	0.65971535	9.816524
73	1.3001	0.65971535	9.816524
74	1.2901	0.65971535	9.816524
75	1.286	0.65971535	9.816524
76	1.2881	0.65971535	9.816524
77	1.2766	0.7067742	10.420017
78	1.2772	0.5331601	9.760937
79	1.2779	0.5331601	9.760937
80	1.2829	0.5331601	9.760937
81	1.2764	0.5331601	9.760937
82	1.2709	0.5331601	9.760937
83	1.2779	0.5331601	9.760937
84	1.2701	0.5331601	9.760937
85	1.2811	0.5331601	9.760937
86	1.2712	0.5331601	9.760937
87	1.2623	0.5331601	9.760937
88	1.2643	0.5331601	9.760937
89	1.2637	0.5331601	9.760937
90	1.2568	0.5331601	9.760937
91	1.2582	0.5331601	9.760937
92	1.2617	0.5331601	9.760937
93	1.2594	0.5331601	9.760937
94	1.2559	0.5331601	9.760937
95	1.2515	0.5331601	9.760937
96	1.2542	0.5331601	9.760937
97	1.2519	0.5331601	9.760937
98	1.2549	0.5331601	9.760937
99	1.2452	0.5331601	9.760937
100	1.2454	0.5331601	9.760937
101	1.2459	0.5331601	9.760937
102	1.214	0.5807701	10.2644415
103	1.216	0.5807701	10.2644415
104	1.2151	0.5807701	10.2644415
105	1.2133	0.5807701	10.2644415
106	1.2145	0.5807701	10.2644415
107	1.2157	0.5807701	10.2644415
108	1.2146	0.5807701	10.2644415
109	1.2157	0.5807701	10.2644415
110	1.2165	0.5807701	10.2644415
111	1.2161	0.5807701	10.2644415
112	1.2158	0.5807701	10.2644415
113	1.2155	0.5807701	10.2644415
114	1.2165	0.5807701	10.2644415
115	1.2166	0.5807701	10.2644415
116	1.2146	0.5807701	10.2644415
117	1.2157	0.5807701	10.2644415
118	1.2166	0.5807701	10.2644415
119	1.2158	0.5807701	10.2644415
120	1.2166	0.5807701	10.2644415
}\ten
	\pgfplotstableread{
epoch	loss	latency	hw
1	1.6449	1.778274059	30.47757149
2	1.6277	3.194540977	50.9011879
3	1.6375	2.712779522	48.53682327
4	1.6141	3.106574059	49.60157394
5	1.6042	3.191408396	49.55162048
6	1.605	3.128895283	49.88319397
7	1.5906	3.252643585	50.41358566
8	1.5933	3.252643585	50.41358566
9	1.5873	2.617302418	46.84857178
10	1.591	3.112787724	49.76491547
11	1.5709	2.818427324	48.29468155
12	1.5606	2.63373208	47.63443756
13	1.552	3.08210206	49.99897766
14	1.5562	2.5927248	47.73401642
15	1.5482	3.035455227	49.35532761
16	1.5454	2.592392683	46.60677719
17	1.5353	2.588470697	47.75377655
18	1.5283	2.589485168	46.55395126
19	1.5328	2.598480463	46.75260544
20	1.5098	2.578891516	46.65887451
21	1.4927	2.642436981	47.81230927
22	1.4854	2.642436981	47.81230927
23	1.4826	2.611551285	48.37950897
24	1.4547	2.611551285	48.37950897
25	1.465	2.611551285	48.37950897
26	1.4484	2.611551285	48.37950897
27	1.437	2.611551285	48.37950897
28	1.4303	2.611551285	48.37950897
29	1.4188	2.611551285	48.37950897
30	1.4147	2.611551285	48.37950897
31	1.4196	2.617227554	48.49504471
32	1.4115	2.617227554	48.49504471
33	1.3855	2.617227554	48.49504471
34	1.391	2.617227554	48.49504471
35	1.3755	2.617227554	48.49504471
36	1.3569	2.58654213	47.5564003
37	1.3242	2.58654213	47.5564003
38	1.3402	2.58654213	47.5564003
39	1.3187	2.58654213	47.5564003
40	1.3247	2.58654213	47.5564003
41	1.2991	2.579116583	47.17623138
42	1.2802	2.579116583	47.17623138
43	1.277	2.579116583	47.17623138
44	1.261	2.598209381	47.83850098
45	1.2658	2.605523109	48.09432983
46	1.2558	2.594461441	47.68046951
47	1.2602	2.594461441	47.68046951
48	1.2463	2.586030483	47.29677963
49	1.2438	2.5788486	46.90988159
50	1.2556	2.160984993	36.99485016
51	1.329	1.51380825	23.0909214
52	1.357	1.111899137	17.98353958
53	1.3864	0.8592139482	14.21051311
54	1.3885	0.7693003416	11.69312096
55	1.4142	0.5537737012	9.044219971
56	1.3984	0.5537737012	9.044219971
57	1.3808	0.5537737012	9.044219971
58	1.3663	0.5537737012	9.044219971
59	1.3665	0.568859458	9.228201866
60	1.3553	0.5537737012	9.044219971
61	1.3463	0.5537737012	9.044219971
62	1.3612	0.5537737012	9.044219971
63	1.3554	0.5537737012	9.044219971
64	1.3471	0.5537737012	9.044219971
65	1.341	0.5537737012	9.044219971
66	1.3401	0.5537737012	9.044219971
67	1.3218	0.5537737012	9.044219971
68	1.3139	0.5537737012	9.044219971
69	1.3049	0.5537737012	9.044219971
70	1.3021	0.5537737012	9.044219971
71	1.2875	0.5568063855	9.10406208
72	1.2784	0.5568063855	9.10406208
73	1.2695	0.5568063855	9.10406208
74	1.2758	0.5568063855	9.10406208
75	1.2755	0.5568063855	9.10406208
76	1.2819	0.5537737012	9.044219971
77	1.2826	0.5537737012	9.044219971
78	1.2702	0.5537737012	9.044219971
79	1.2631	0.5537737012	9.044219971
80	1.2659	0.5537737012	9.044219971
81	1.2556	0.5537737012	9.044219971
82	1.2566	0.5537737012	9.044219971
83	1.2583	0.5537737012	9.044219971
84	1.2563	0.5537737012	9.044219971
85	1.2456	0.5625766516	9.19591713
86	1.2451	0.5625766516	9.19591713
87	1.2444	0.5625766516	9.19591713
88	1.244	0.5625766516	9.19591713
89	1.2282	0.5652542114	9.250875473
90	1.2224	0.5652542114	9.250875473
91	1.2215	0.5652542114	9.250875473
92	1.225	0.5652542114	9.250875473
93	1.2221	0.5652542114	9.250875473
94	1.2146	0.5652542114	9.250875473
95	1.2122	0.5652542114	9.250875473
96	1.2166	0.5652542114	9.250875473
97	1.2154	0.5652542114	9.250875473
98	1.2176	0.5652542114	9.250875473
99	1.2109	0.5652542114	9.250875473
100	1.2113	0.5652542114	9.250875473
101	1.2111	0.5652542114	9.250875473
102	1.2089	0.5652542114	9.250875473
103	1.2091	0.5652542114	9.250875473
104	1.2073	0.5652542114	9.250875473
105	1.2067	0.5652542114	9.250875473
106	1.2089	0.5652542114	9.250875473
107	1.2089	0.5652542114	9.250875473
108	1.2074	0.5652542114	9.250875473
109	1.2057	0.5652542114	9.250875473
110	1.2074	0.5652542114	9.250875473
111	1.2083	0.5652542114	9.250875473
112	1.2075	0.5652542114	9.250875473
113	1.2076	0.5652542114	9.250875473
114	1.2077	0.5652542114	9.250875473
115	1.2078	0.5652542114	9.250875473
116	1.2068	0.5652542114	9.250875473
117	1.2075	0.5652542114	9.250875473
118	1.2072	0.5652542114	9.250875473
119	1.207	0.5652542114	9.250875473
120	1.2086	0.5652542114	9.250875473
}\seven

	\pgfplotstableread{
epoch	loss	latency	hw
1	1.6297	1.778274059	30.47757149
2	1.6133	2.593407869	46.47140884
3	1.6112	2.647926569	47.62445068
4	1.6064	3.281603813	50.84550095
5	1.5836	3.248112679	50.34378815
6	1.5722	3.124143124	49.84408188
7	1.5721	3.123623133	49.83988953
8	1.5844	3.123623133	49.83988953
9	1.5557	3.089301825	50.06228256
10	1.5619	3.057292461	49.59770966
11	1.5449	2.606035233	47.94458771
12	1.5564	2.606035233	47.94458771
13	1.5469	2.545256615	46.18984604
14	1.5276	3.053162575	49.54432678
15	1.503	2.637943745	48.4595871
16	1.4939	2.613802433	48.41640854
17	1.4978	2.633444071	47.30570984
18	1.4918	2.633444071	47.30570984
19	1.4733	2.633444071	47.30570984
20	1.4805	2.633444071	47.30570984
21	1.46	2.633444071	47.30570984
22	1.455	2.635944128	47.88682938
23	1.4311	2.635944128	47.88682938
24	1.4133	2.635944128	47.88682938
25	1.4052	2.633444071	47.30570984
26	1.393	2.635944128	47.88682938
27	1.3854	2.610327721	46.88931274
28	1.3752	2.633444071	47.30570984
29	1.3483	2.633444071	47.30570984
30	1.3587	2.650255442	48.14127731
31	1.3441	2.626600027	47.72047043
32	1.3435	2.626600027	47.72047043
33	1.3222	2.626600027	47.72047043
34	1.3155	2.626600027	47.72047043
35	1.3154	2.624819756	47.15235138
36	1.3083	2.626600027	47.72047043
37	1.3023	2.626600027	47.72047043
38	1.3007	2.626600027	47.72047043
39	1.2772	2.626600027	47.72047043
40	1.2953	2.626600027	47.72047043
41	1.2756	2.626600027	47.72047043
42	1.2678	2.626600027	47.72047043
43	1.2631	2.626600027	47.72047043
44	1.2458	2.624819756	47.15235138
45	1.2559	2.624819756	47.15235138
46	1.2299	2.624819756	47.15235138
47	1.2272	2.624819756	47.15235138
48	1.2125	2.624819756	47.15235138
49	1.2049	2.624819756	47.15235138
50	1.2025	2.57242775	46.61846542
51	1.1936	2.57242775	46.61846542
52	1.1847	2.557023048	46.22537994
53	1.1766	2.557023048	46.22537994
54	1.1705	2.060461044	62.02939987
55	1.1553	2.037582874	61.44921875
56	1.1569	1.952775002	60.07381439
57	1.1821	2.090073347	34.64238358
58	1.2082	1.651957512	23.15379143
59	1.2192	1.517073274	21.13451576
60	1.2237	1.517377853	20.70830917
61	1.3547	1.186943769	16.81595993
62	1.3513	1.074766874	14.67942238
63	1.3431	0.8884114623	12.34180069
64	1.3673	0.829405725	12.24864864
65	1.3873	0.7831580639	11.6617527
66	1.3928	0.6927828789	10.26866055
67	1.4179	0.523512125	8.230183601
68	1.3992	0.523512125	8.230183601
69	1.3933	0.523512125	8.230183601
70	1.3736	0.523512125	8.230183601
71	1.3698	0.523512125	8.230183601
72	1.3656	0.523512125	8.230183601
73	1.3659	0.523512125	8.230183601
74	1.3531	0.523512125	8.230183601
75	1.3531	0.523512125	8.230183601
76	1.3495	0.523512125	8.230183601
77	1.3423	0.523512125	8.230183601
78	1.3307	0.523512125	8.230183601
79	1.3388	0.523512125	8.230183601
80	1.3388	0.523512125	8.230183601
81	1.3341	0.523512125	8.230183601
82	1.3399	0.523512125	8.230183601
83	1.3342	0.523512125	8.230183601
84	1.3315	0.523512125	8.230183601
85	1.3288	0.523512125	8.230183601
86	1.3247	0.523512125	8.230183601
87	1.3181	0.523512125	8.230183601
88	1.3196	0.523512125	8.230183601
89	1.3171	0.523512125	8.230183601
90	1.3136	0.523512125	8.230183601
91	1.3088	0.523512125	8.230183601
92	1.3121	0.523512125	8.230183601
93	1.3092	0.523512125	8.230183601
94	1.3094	0.523512125	8.230183601
95	1.3048	0.523512125	8.230183601
96	1.3036	0.523512125	8.230183601
97	1.3056	0.523512125	8.230183601
98	1.3046	0.523512125	8.230183601
99	1.3016	0.523512125	8.230183601
100	1.298	0.523512125	8.230183601
101	1.3258	0.4879005253	7.75139904
102	1.3251	0.4879005253	7.75139904
103	1.2963	0.523512125	8.230183601
104	1.2961	0.523512125	8.230183601
105	1.2963	0.523512125	8.230183601
106	1.2964	0.523512125	8.230183601
107	1.2981	0.523512125	8.230183601
108	1.2988	0.523512125	8.230183601
109	1.3222	0.4879005253	7.75139904
110	1.325	0.4879005253	7.75139904
111	1.3244	0.4879005253	7.75139904
112	1.3242	0.4879005253	7.75139904
113	1.3238	0.4879005253	7.75139904
114	1.3244	0.4879005253	7.75139904
115	1.3245	0.4879005253	7.75139904
116	1.3229	0.4879005253	7.75139904
117	1.3238	0.4879005253	7.75139904
118	1.3239	0.4879005253	7.75139904
119	1.3235	0.4879005253	7.75139904
120	1.3252	0.4879005253	7.75139904
}\six
	\pgfplotstableread{
epoch	loss	latency	hw
1	1.6578	1.778274059	30.47757149
2	1.6427	2.65216589	47.57966614
3	1.6294	3.114024639	50.49303055
4	1.6309	2.579284191	46.1690979
5	1.6173	3.036882162	48.70802689
6	1.597	3.036882162	48.70802689
7	1.5914	3.042901039	48.92520905
8	1.6027	3.042901039	48.92520905
9	1.5944	3.04500699	48.82202148
10	1.5803	3.047152042	49.03587341
11	1.6093	3.047152042	49.03587341
12	1.5819	3.031380892	48.80222702
13	1.5916	2.583276272	47.28871155
14	1.5756	2.583276272	47.28871155
15	1.5704	2.583276272	47.28871155
16	1.5457	2.58577776	47.724823
17	1.5422	2.578863621	47.01396561
18	1.5362	2.578863621	47.01396561
19	1.5276	2.578863621	47.01396561
20	1.513	2.588037968	47.48717117
21	1.5045	2.588037968	47.48717117
22	1.4886	2.588037968	47.48717117
23	1.4864	2.588037968	47.48717117
24	1.4734	2.592302084	47.7577095
25	1.458	2.592302084	47.7577095
26	1.4611	2.592302084	47.7577095
27	1.4323	2.592302084	47.7577095
28	1.439	2.592302084	47.7577095
29	1.4124	2.592302084	47.7577095
30	1.4287	2.613755703	48.30154037
31	1.4165	2.613755703	48.30154037
32	1.4085	2.613755703	48.30154037
33	1.4086	2.613755703	48.30154037
34	1.3859	2.613755703	48.30154037
35	1.3766	2.613755703	48.30154037
36	1.3602	2.613755703	48.30154037
37	1.3604	2.613755703	48.30154037
38	1.3437	2.613755703	48.30154037
39	1.322	2.613755703	48.30154037
40	1.3477	2.613755703	48.30154037
41	1.3156	2.613755703	48.30154037
42	1.2998	2.613755703	48.30154037
43	1.3066	2.617639542	48.45793152
44	1.284	2.617639542	48.45793152
45	1.2851	2.617639542	48.45793152
46	1.2706	2.617639542	48.45793152
47	1.268	2.617639542	48.45793152
48	1.245	2.617639542	48.45793152
49	1.2438	2.617639542	48.45793152
50	1.2444	2.617639542	48.45793152
51	1.2353	2.617639542	48.45793152
52	1.2207	2.617639542	48.45793152
53	1.2227	2.617639542	48.45793152
54	1.2242	2.617639542	48.45793152
55	1.2068	2.617639542	48.45793152
56	1.2024	2.617639542	48.45793152
57	1.1882	2.617639542	48.45793152
58	1.1901	2.617639542	48.45793152
59	1.196	2.677917957	49.17676163
60	1.1935	2.617639542	48.45793152
61	1.164	2.617639542	48.45793152
62	1.177	2.617639542	48.45793152
63	1.1681	2.617639542	48.45793152
64	1.1594	2.600459576	47.95906448
65	1.1664	2.594027042	47.70052719
66	1.1636	2.584893227	47.3110733
67	1.1659	2.600597143	47.80330658
68	1.1683	2.584893227	47.3110733
69	1.1526	2.584893227	47.3110733
70	1.162	2.530063152	46.27986145
71	1.1743	1.795839667	31.32464218
72	1.2728	0.8387854099	13.60916519
73	1.3009	0.7853357196	13.04389477
74	1.2886	0.5968230367	10.32444668
75	1.301	0.5985590219	10.08871555
76	1.3047	0.5985590219	10.08871555
77	1.2966	0.5968230367	10.32444668
78	1.2908	0.5985590219	10.08871555
79	1.2833	0.5985590219	10.08871555
80	1.2815	0.5985590219	10.08871555
81	1.2817	0.5985590219	10.08871555
82	1.2811	0.644318223	10.99305344
83	1.2666	0.644318223	10.99305344
84	1.2627	0.5985590219	10.08871555
85	1.2556	0.644318223	10.99305344
86	1.2569	0.5985590219	10.08871555
87	1.2541	0.5985590219	10.08871555
88	1.2576	0.5985590219	10.08871555
89	1.2538	0.5985590219	10.08871555
90	1.25	0.5968230367	10.32444668
91	1.2604	0.6634319425	10.24217415
92	1.2549	0.5985590219	10.08871555
93	1.2499	0.5985590219	10.08871555
94	1.246	0.5985590219	10.08871555
95	1.2441	0.644318223	10.99305344
96	1.2402	0.5985590219	10.08871555
97	1.2375	0.644318223	10.99305344
98	1.2351	0.644318223	10.99305344
99	1.2313	0.644318223	10.99305344
100	1.235	0.6634319425	10.24217415
101	1.2315	0.5985590219	10.08871555
102	1.2325	0.6634319425	10.24217415
103	1.229	0.5985590219	10.08871555
104	1.2293	0.6634319425	10.24217415
105	1.2299	0.644318223	10.99305344
106	1.2293	0.5985590219	10.08871555
107	1.229	0.5985590219	10.08871555
108	1.2283	0.6634319425	10.24217415
109	1.2275	0.5985590219	10.08871555
110	1.2283	0.5985590219	10.08871555
111	1.2269	0.5985590219	10.08871555
112	1.2374	0.6027772427	10.55319023
113	1.236	0.6027772427	10.55319023
114	1.237	0.6027772427	10.55319023
115	1.2379	0.6027772427	10.55319023
116	1.2371	0.6027772427	10.55319023
117	1.2378	0.6027772427	10.55319023
118	1.2379	0.6027772427	10.55319023
119	1.2382	0.6027772427	10.55319023
120	1.2389	0.6027772427	10.55319023
}\five	
	\pgfplotstableread{
epoch	loss	latency	hw
1	1.6186	2.1226048	31.667557
2	1.603	3.036257	47.22694
3	1.6134	3.0339904	47.230118
4	1.5948	3.046185	47.447964
5	1.5943	3.046185	47.447964
6	1.6003	3.0506997	47.507977
7	1.6077	3.0445912	47.355705
8	1.5938	3.0506997	47.507977
9	1.5661	3.0216923	47.008316
10	1.5649	3.0216923	47.008316
11	1.5592	3.0345888	47.191425
12	1.5507	3.0345888	47.191425
13	1.5416	3.0586147	47.577488
14	1.5314	3.0586147	47.577488
15	1.5281	3.0733907	47.877792
16	1.5284	3.0733907	47.877792
17	1.512	3.103363	48.31974
18	1.498	3.103363	48.31974
19	1.5127	2.9709084	47.745132
20	1.4943	2.9709084	47.745132
21	1.511	2.7500381	46.765343
22	1.4896	2.7500381	46.765343
23	1.4735	2.7500381	46.765343
24	1.4571	2.7500381	46.765343
25	1.4528	2.5751612	47.04934
26	1.4506	2.5751612	47.04934
27	1.4448	2.5751612	47.04934
28	1.4334	2.5751612	47.04934
29	1.413	2.5672038	47.065754
30	1.4205	2.571679	46.860245
31	1.3961	2.571679	46.860245
32	1.393	2.5751612	47.04934
33	1.4035	2.5751612	47.04934
34	1.3907	2.5751612	47.04934
35	1.3762	2.5751612	47.04934
36	1.3715	2.5751612	47.04934
37	1.3483	2.5751612	47.04934
38	1.3469	2.571679	46.860245
39	1.3308	2.5653968	46.658104
40	1.3235	2.5653968	46.658104
41	1.3104	2.571679	46.860245
42	1.2974	2.571679	46.860245
43	1.2936	2.571679	46.860245
44	1.2776	2.5795138	47.264606
45	1.2863	2.5795138	47.264606
46	1.2623	2.5795138	47.264606
47	1.2712	2.5795138	47.264606
48	1.2462	2.5795138	47.264606
49	1.224	2.5860343	47.636005
50	1.2205	2.5860343	47.636005
51	1.2123	2.5860343	47.636005
52	1.2138	2.5860343	47.636005
53	1.2033	2.6115885	48.14092
54	1.2155	2.6115885	48.14092
55	1.2062	2.6115885	48.14092
56	1.1874	2.6115885	48.14092
57	1.1938	2.6115885	48.14092
58	1.1779	2.6115885	48.14092
59	1.165	2.6115885	48.14092
60	1.1689	2.6115885	48.14092
61	1.1473	2.6214297	48.326347
62	1.1491	2.6214297	48.326347
63	1.1489	2.6214297	48.326347
64	1.1386	2.6861026	49.045513
65	1.1442	2.700948	49.282227
66	1.1254	2.700948	49.282227
67	1.1321	2.700948	49.282227
68	1.1248	2.6362162	48.566544
69	1.1191	2.6362162	48.566544
70	1.1134	2.6362162	48.566544
71	1.1192	2.6362162	48.566544
72	1.1081	2.6362162	48.566544
73	1.1035	3.2679143	50.64346
74	1.1002	2.6362162	48.566544
75	1.0979	2.6362162	48.566544
76	1.089	2.6362162	48.566544
77	1.0985	2.6362162	48.566544
78	1.0875	2.6362162	48.566544
79	1.0941	2.6861026	49.045513
80	1.0954	2.6214297	48.326347
81	1.097	2.5899062	47.781784
82	1.0941	2.5852954	47.582047
83	1.0946	2.5410228	46.97964
84	1.1208	2.1318498	37.14384
85	1.1364	2.0933192	36.611465
86	1.16	1.9482007	34.25346
87	1.1803	1.7093284	30.647263
88	1.1861	1.3611169	24.3079
89	1.2862	0.87240505	16.86637
90	1.2932	0.6812021	12.043222
91	1.2984	0.6760671	11.813834
92	1.312	0.6266271	11.473738
93	1.3058	0.6266271	11.473738
94	1.3023	0.6266271	11.473738
95	1.2884	0.62843907	11.301217
96	1.2829	0.62843907	11.301217
97	1.2928	0.62843907	11.301217
98	1.2857	0.62843907	11.301217
99	1.2792	0.62843907	11.301217
100	1.2804	0.62843907	11.301217
101	1.2729	0.62843907	11.301217
102	1.2762	0.6298126	11.495372
103	1.2745	0.62843907	11.301217
104	1.2698	0.62843907	11.301217
105	1.2665	0.62843907	11.301217
106	1.2685	0.62843907	11.301217
107	1.2717	0.62843907	11.301217
108	1.2749	0.6298126	11.495372
109	1.2713	0.62843907	11.301217
110	1.2705	0.62843907	11.301217
111	1.2654	0.63919777	11.5891075
112	1.2652	0.63919777	11.5891075
113	1.2769	0.5730201	11.170047
114	1.2773	0.5730201	11.170047
115	1.2791	0.5730201	11.170047
116	1.2713	0.58450156	10.821378
117	1.2715	0.58450156	10.821378
118	1.2631	0.6253315	11.2674885
119	1.2627	0.6256592	11.420601
120	1.2737	0.6256592	11.420601
}\four
	\pgfplotstableread{
epoch	loss	latency	hw
1	1.6143	2.122604847	31.66755676
2	1.6013	3.105695724	48.3053093
3	1.625	3.190420628	49.48879242
4	1.6149	3.213324547	49.82917786
5	1.6059	3.23784256	50.1993866
6	1.608	3.228127003	50.08465958
7	1.5805	3.228127003	50.08465958
8	1.6003	3.238307714	50.22866821
9	1.5718	3.072970867	47.95578384
10	1.5467	3.073261023	47.91960526
11	1.5569	3.07943821	48.01018906
12	1.5698	3.244812965	50.27252197
13	1.561	3.238474607	50.16415405
14	1.5446	3.244812965	50.27252197
15	1.5297	3.244812965	50.27252197
16	1.5209	3.2498703	50.36527252
17	1.5246	3.2498703	50.36527252
18	1.5209	3.2498703	50.36527252
19	1.4941	3.241446495	50.24150085
20	1.487	3.241446495	50.24150085
21	1.4897	3.257283688	50.47384644
22	1.4819	3.257283688	50.47384644
23	1.4556	3.257283688	50.47384644
24	1.4393	3.257283688	50.47384644
25	1.4366	3.257283688	50.47384644
26	1.4463	2.838953733	48.48278427
27	1.4225	3.257283688	50.47384644
28	1.4187	2.838953733	48.48278427
29	1.4041	2.838953733	48.48278427
30	1.4084	3.257283688	50.47384644
31	1.3908	2.859630346	48.7885704
32	1.3902	2.859630346	48.7885704
33	1.3729	2.859630346	48.7885704
34	1.3684	2.859630346	48.7885704
35	1.3622	2.859630346	48.7885704
36	1.3555	2.859630346	48.7885704
37	1.3375	2.859630346	48.7885704
38	1.3217	2.848221779	48.65361404
39	1.308	2.859630346	48.7885704
40	1.3013	2.859630346	48.7885704
41	1.2985	2.859630346	48.7885704
42	1.2973	2.859630346	48.7885704
43	1.29	2.859630346	48.7885704
44	1.2743	2.859630346	48.7885704
45	1.274	2.859630346	48.7885704
46	1.2537	2.859630346	48.7885704
47	1.2545	2.859630346	48.7885704
48	1.2274	2.830329418	49.50868225
49	1.2224	2.830329418	49.50868225
50	1.2325	2.830329418	49.50868225
51	1.2066	2.830329418	49.50868225
52	1.2189	2.859630346	48.7885704
53	1.1937	2.830329418	49.50868225
54	1.2147	2.830329418	49.50868225
55	1.1759	2.830329418	49.50868225
56	1.1788	2.830329418	49.50868225
57	1.1882	2.830329418	49.50868225
58	1.1681	2.830329418	49.50868225
59	1.1706	2.830329418	49.50868225
60	1.1625	2.830329418	49.50868225
61	1.1402	2.859630346	48.7885704
62	1.1518	2.859630346	48.7885704
63	1.1291	2.859630346	48.7885704
64	1.1343	2.702782154	49.33873367
65	1.1202	2.632526875	48.58271027
66	1.1177	2.632526875	48.58271027
67	1.1248	2.632526875	48.58271027
68	1.1127	2.632526875	48.58271027
69	1.0987	2.632526875	48.58271027
70	1.0989	2.632526875	48.58271027
71	1.0937	2.632526875	48.58271027
72	1.0858	2.632526875	48.58271027
73	1.081	2.632526875	48.58271027
74	1.0836	2.632526875	48.58271027
75	1.0889	2.632526875	48.58271027
76	1.0743	2.632526875	48.58271027
77	1.0707	2.632526875	48.58271027
78	1.0632	2.632526875	48.58271027
79	1.0631	2.632526875	48.58271027
80	1.0657	2.632526875	48.58271027
81	1.06	2.632526875	48.58271027
82	1.0495	2.632526875	48.58271027
83	1.0491	2.632526875	48.58271027
84	1.0459	3.118922949	50.42047119
85	1.0491	3.118922949	50.42047119
86	1.0446	3.118922949	50.42047119
87	1.0434	3.118922949	50.42047119
88	1.0414	3.118922949	50.42047119
89	1.0355	2.632526875	48.58271027
90	1.0334	2.632526875	48.58271027
91	1.0374	2.632526875	48.58271027
92	1.0322	2.632526875	48.58271027
93	1.033	2.632526875	48.58271027
94	1.0243	2.632526875	48.58271027
95	1.0321	2.632526875	48.58271027
96	1.023	2.632526875	48.58271027
97	1.0252	2.632526875	48.58271027
98	1.0268	2.632526875	48.58271027
99	1.0161	2.632526875	48.58271027
100	1.0279	2.632526875	48.58271027
101	1.0218	2.632526875	48.58271027
102	1.0225	2.632526875	48.58271027
103	1.0208	2.632526875	48.58271027
104	1.0188	2.632526875	48.58271027
105	1.0176	2.632526875	48.58271027
106	1.0189	2.632526875	48.58271027
107	1.0184	2.632526875	48.58271027
108	1.0187	2.632526875	48.58271027
109	1.0195	2.59595108	48.0001297
110	1.0208	2.59595108	48.0001297
111	1.02	2.59595108	48.0001297
112	1.018	2.60438776	47.84963226
113	1.0139	2.609182119	48.00770569
114	1.0154	2.609182119	48.00770569
115	1.016	2.609182119	48.00770569
116	1.047	2.222016811	37.95551682
117	1.0471	2.091850042	38.29021454
118	1.0563	1.925158501	35.00986862
119	1.0877	1.801559448	31.40759087
120	1.1095	1.801559448	31.40759087
}\three	
	\pgfplotstableread{
epoch	loss	latency	hw

}\two	
	\pgfplotstableread{
epoch	loss	latency	hw

}\one	

	\pgfplotstableread{
epoch	loss	latency	hw
-100	  0 0.666 0
400	0 0.66 0
}\constraint
\pgfplotstableread{

epoch	latency
1	-1
2	-1
3	-1
4	-1
5	-1
6	-1
7	-1
8	-1
9	-1
10	-1
}\dummy		